\documentclass[runningheads]{llncs}

\usepackage{eccv}

\usepackage{eccvabbrv}

\usepackage{graphicx}
\usepackage{booktabs}

\usepackage[accsupp]{axessibility}  

\usepackage{hyperref}

\usepackage{orcidlink}

\usepackage{algorithm}
\usepackage{algorithmic}
\usepackage{graphicx}
\usepackage{amsmath}
\usepackage{amssymb}
\usepackage{subcaption}
\usepackage{bbold}
\usepackage{array}
\usepackage{multirow}
\newlength\myindent
\newcommand\bindent{%
  \begingroup
  \setlength{\itemindent}{\myindent}
  \addtolength{\algorithmicindent}{\myindent}
}
\newcommand\eindent{\endgroup}

\begin{document}

\title{FedHide: Federated Learning by Hiding in the Neighbors}


\author{Hyunsin Park\orcidlink{0000-0003-3556-5792} \and
Sungrack Yun\orcidlink{0000-0003-2462-3854}}

\authorrunning{H.~Park~and~S.~Yun}

\institute{Qualcomm AI Research$^{\dagger}$ \\
\email{\{hyunsinp,sungrack\}@qti.qualcomm.com}}

\maketitle

\begin{abstract}
We propose a prototype-based federated learning method designed for embedding networks in classification or verification tasks. Our focus is on scenarios where each client has data from a single class. The main challenge is to develop an embedding network that can distinguish between different classes while adhering to privacy constraints. Sharing true class prototypes with the server or other clients could potentially compromise sensitive information. To tackle this issue, we propose a proxy class prototype that will be shared among clients instead of the true class prototype. Our approach generates proxy class prototypes by linearly combining them with their nearest neighbors. This technique conceals the true class prototype while enabling clients to learn discriminative embedding networks. We compare our method to alternative techniques, such as adding random Gaussian noise and using random selection with cosine similarity constraints. Furthermore, we evaluate the robustness of our approach against gradient inversion attacks and introduce a measure for prototype leakage. This measure quantifies the extent of private information revealed when sharing the proposed proxy class prototype. Moreover, we provide a theoretical analysis of the convergence properties of our approach. Our proposed method for federated learning from scratch demonstrates its effectiveness through empirical results on three benchmark datasets: CIFAR-100, VoxCeleb1, and VGGFace2.
\keywords{Federated learning \and Contrastive learning \and User verification}
\end{abstract}

\section{Introduction}
{\let\thefootnote\relax\footnotetext{{
$\dagger$ Qualcomm AI Research is an initiative of Qualcomm Technologies, Inc.}}}
The problem of training embedding networks has been widely studied due to its applicability in various tasks, such as identification, verification, retrieval, and clustering \cite{snyder2017deep, yun2019end, wang2018AMS_fv, cao2018automated, nguyen2017iris, schroff2015facenet, mikolov2013distributed}. 
These networks are typically trained using a loss function that simultaneously minimizes the distance between instance embeddings belonging to the same class and maximizes the distance between instance embeddings from different classes. In recent years, deep neural networks trained on large datasets have been employed to obtain nonlinear embeddings \cite{zhong2019adversarial,guan2020collaborative,ge2017improving,chung2018voxceleb2}. However, collecting large and high-quality data for training deep networks remains expensive for real-world applications \cite{zhao2017weakly,zhu2017deep,zhang2020sce}. 

One approach to address the data collection problem is to train the model using a federated learning framework.
In this framework, a global model is iteratively updated by aggregating local models without requiring direct access to local data~\cite{mcmahan2017learning,bonawitz2019towards,konevcny2016federated,sattler2019robust,karimireddy2020scaffold}. 
Specifically, we consider a scenario where each client has access to data from only one target class and cannot share embeddings with the server or other clients.
In such a setting, it becomes challenging for each client to learn an embedding network that discriminates different classes in the embedding space due to the lack of information about other clients' class prototypes. Consequently, the learned class prototypes might collapse into a single embedding. 

The problem of training embedding networks in a federated setup has been recently explored in various settings.  
Federated Averaging with Spreadout (FedAwS) \cite{yu2020federated} learns an embedding network for multi-class classification in the federated setup, where each client has access to only positive labels. In this method, client embeddings are shared with the server, and a regularization term is applied to increase pairwise distances between embeddings. 
However, the server is assumed not to share client class prototypes with others.
Unfortunately, adversaries with access to the server may perform a model-inversion attack \cite{fredrikson2015model, geiping2020inverting, huang2021evaluating} to reconstruct inputs using a pretrained model and a target identity related to the class prototype. 
Another recent approach is Federated User Verification (FedUV) \cite{hosseini2021federated}, which proposes to use predefined codewords guaranteeing a minimum distance between class prototypes. Consequently, FedUV does not require sharing class prototypes with the server. However, it does not take into account the similarity of clients' data during training embeddings.

To address this problem, we propose a federated learning framework in which each client updates its local model with a contrastive learning loss to minimize intra-class variance and maximize inter-class variance.
This approach requires sharing class prototypes with other clients, potentially exposing security-sensitive information.
Instead, we introduce a method called FedHide, in which clients share proxy class prototypes generated by linearly combining them with their nearest neighbors to reduce the expose of security-sensitive information.
We also provide a theoretical analysis of the convergence rate of FedHide when dealing with non-convex objectives.
Empirically, our approach reduces the exposure of sensitive embeddings to other users while maintaining discriminating performance across datasets such as CIFAR-100, VoxCeleb1, and VGGFace2 datasets.

\section{Related Works}
In this paper, we consider a scenario where each client has access to data from only one class. In such cases, we can naively apply one-class classification approaches, such as DeepSVDD (Deep Support Vector Data Description) \cite{ruff2018deep} and DROCC (Deep Robust One-Class Classification) \cite{goyal2020drocc}.
In DeepSVDD, it trains an embedding network by minimizing the volume of a hypersphere that encloses the instance embeddings of the data. 
By minimizing the hypersphere’s volume, the network extracts common factors of variation, aiming to closely map data points to the center of the hypersphere. 
To prevent hypersphere collapse, DeepSVDD uses neural networks without bias terms or bounded activation functions.
Motivated by the observation that data from special classes lie on a low-dimensional manifold, DROCC introduces a discriminative component. This component generates anomalous data, which are then used to train the embedding network.
However, the focus of this paper lies in finding a way to utilize other clients’ information without compromising privacy.

Federated learning (FL) is a method for training a model across distributed edge (client) devices without sharing local data information.
In each round of the learning process, the server broadcasts the current global model to selected clients.
After clients update their local models from the shared global model using local data, these local models are uploaded to the server. 
Finally, the server aggregates the local models to update the global model.
A popular FL algorithm is Federated Averaging (FedAvg) \cite{mcmahan2017learning}. 
However, in our scenario where each client has access to data from only one class, sharing the parameters of the output layer, called a class prototype, with other clients is inappropriate. The class prototype contains client-specific information and could be exploited in a gradient inversion attack \cite{geiping2020inverting,  huang2021evaluating}. Such an attack reconstructs an input by minimizing the discrepancy between the gradient of a reconstructed input image and the gradient uploaded from a client. It highlights that sharing gradients does not guarantee client privacy within the federated learning framework. We will demonstrate the robustness of our method against this attack.

There are FL methods that focus on solving problems where data are non-identically distributed among clients. 
In FedProx \cite{li2020federated_FedProx}, local updates are constrained by the $L^2$-norm distance.
SCAFFOLD \cite{karimireddy2020scaffold} corrects local updates via variance reduction.
MOON \cite{li2021model} is a model-level contrastive FL method that corrects local updates by maximizing the agreement of representation learned by the current local model and the representation learned by the global model. 
However, these methods do not specifically address our scenario of having a single class per client, which represents an extremely non-iid case.

There are several works to handle the extremely non-iid case. 

FedAwS \cite{yu2020federated} trains an embedding network for multi-class classification in the federated setting. Each client has access to only positive data. The loss function of FedAwS is based on a contrastive loss, aiming to minimize intra-class variance while simultaneously maximizing inter-class variance. At each client, a similar loss used in DeepSVDD is optimized to train a local model. Each client then uploads its local model and a class prototype to the server. Instead of directly sharing the class prototype with other clients, FedAwS optimizes a regularization term to spread out the class prototypes. However, FedAwS still requires sharing class prototypes with the server, which may raise privacy concerns.
FedFace \cite{aggarwal2021fedface} is proposed for collaborative learning of face recognition models based on FedAwS. It shows good performance on face recognition benchmarks. However, it requires a well-pretrained model as an initial global model and still faces privacy leakage issues.
FedUV \cite{hosseini2021federated} aims to eliminate the requirement of sharing class prototypes with the server in federated learning of user verification models. The authors propose using predefined codewords of an error-correcting code as class prototypes. This approach allows clients to collaboratively train user verification models without compromising privacy. However, FedUV has a limitation to model the similarity between clients in the embedding space, as the codewords are predefined without considering local data characteristics.

Our proposed method is related to prototype-based federated learning approaches. 
In FedProto \cite{tan2021fedproto}, each client has a different embedding network and does not share model parameters but only class prototypes. This approach avoids compromising private information.
In FedPCL \cite{tan2022federated}, clients jointly learn to fuse representations generated by multiple fixed pre-trained models using a prototype-wise contrastive learning approach.
FedNH \cite{dai2022tackling} proposes using initial class prototypes uniformly distributed in the latent space and smoothly infusing class information into these prototypes. 
However, they do not maintain global embedding networks, and deploying a global embedding network to unseen clients is not feasible.
ProtoFL \cite{kim2023protofl} is a method designed to enhance the representation power of a global model and reduce communication costs. However, it requires an off-the-shelf model and dataset at the server.

\section{Method}

\subsection{Federated Learning Based on a Contrastive Learning Loss with Proxy Prototypes}

\begin{figure}[t]
    \centering
    \includegraphics[width=\textwidth]{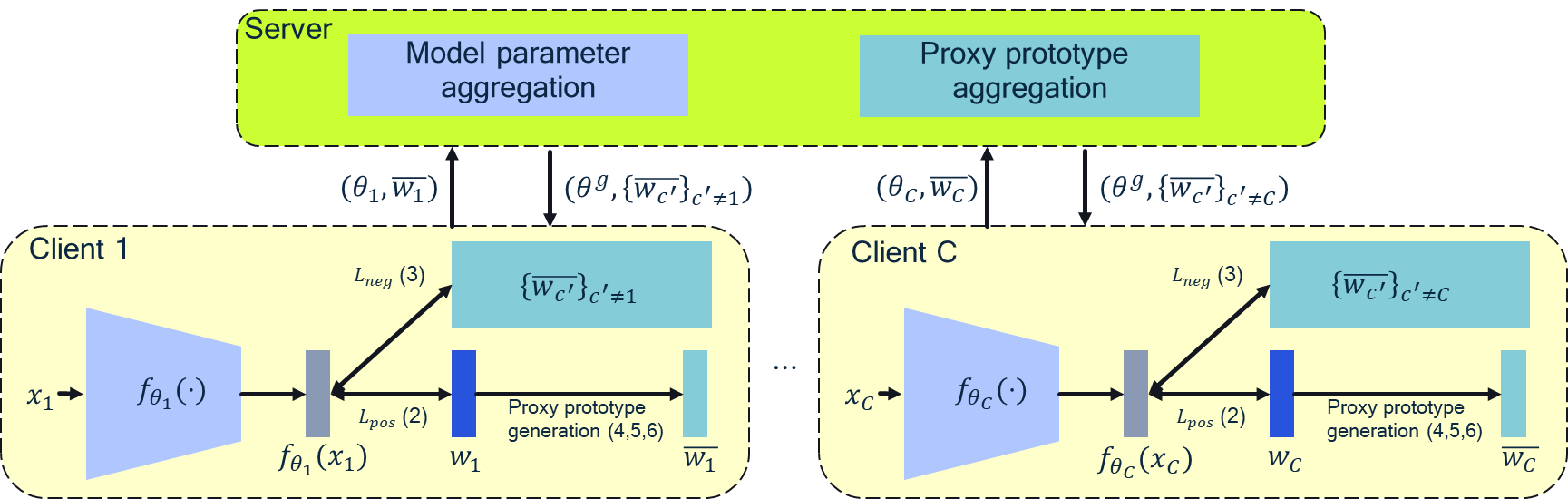}
    \caption{A diagram of the FedHide algorithm. Each client updates its local embedding network and prototype using a contrastive loss and shared proxy prototypes. The server collects the local updates and proxy prototypes, then broadcasts the aggregated model parameters and proxy prototypes to all clients.}
    \label{fig:fedhide_diagram}
\end{figure}

We propose an FL framework in which clients update their local models using a contrastive learning loss to minimize intra-class variance and simultaneously maximize inter-class variance. Instead of sharing the true class prototype that represents the instance embeddings of local data, we share a proxy class prototype. This approach reduces the exposure of security-sensitive information and allows us to learn an embedding network that discriminates between different clients. The server collects the local embedding networks and proxy prototypes, then broadcasts aggregated model parameters and proxy prototypes to all clients. An overview of the proposed framework is shown in Fig. \ref{fig:fedhide_diagram}.

\begin{algorithm}[t]
   \caption{FedHide. The $C$ clients are indexed by $c$, $M$ is the number of clients participated at each round, $E$ is the number of local iterations, and $\eta$ is the local learning rate.}
   \label{main-algo}
\begin{algorithmic}
   \STATE \textbf{Server executes:} 
   \bindent
   \STATE Initialize global model $\theta_0$ and proxy class prototypes $\{\bar{w}_{c,0}\}_{c=1}^{C}$
   \FOR{each round $t = 1, 2, \ldots $} 
        \STATE $S \leftarrow$ (random set of $M$ clients among the $C$ clients)
        \FOR{each client $c \in S$} 
        \STATE $\theta_{c,t}, \bar{w}_{c,t}  \leftarrow $  ClientUpdate(c, $\theta_{t-1}$, $\{\bar{w}_{c',t-1}\}_{c' \neq c}$)
        \ENDFOR        
        \STATE $\theta_t \leftarrow \frac{1}{M}\sum_{c \in S} \theta_{c,t}$    ~~~~~~~~~~~~~~~~ // {\it Global model update by averaging}
        \STATE update proxy class prototypes by replacing  $\{\bar{w}_{c,t-1}\}_{c \in S}$ with $\{\bar{w}_{c,t} \}_{c \in S}$
    \ENDFOR
    \eindent
    \newline
    \STATE \textbf{ClientUpdate}($c, \theta_c, \{\bar{w}_{c'}\}_{c' \neq c}$):
    \bindent
    \FOR{each local iteration $i = 1, 2, \ldots, E$}
        \STATE Update local model and prototype $(\theta_c, w_c)$ by using the main loss of Eq. \ref{general_loss} 
    \ENDFOR
    \STATE Generate a proxy class prototype $\bar{w}_c$ using one of Eq. \ref{FedGN}, \ref{FedCS}, and \ref{FedHide}
    \STATE \textbf{Return} $(\theta_c, \bar{w}_c)$
    \eindent
\end{algorithmic}
\end{algorithm}

Let $x \in \mathcal{X}$ represent an input. An embedding network $f_\theta (\cdot): \mathcal{X} \rightarrow \mathbb{R}^d$ takes $x$ as an input and produces an instance embedding vector $f_\theta(x)$. In this paper, we utilize $L^2$-normalized instance embeddings and class prototypes.
The proposed method relies on the following loss functions, including a positive loss and a negative loss for the $c$-th client, 
\begin{equation} \label{general_loss}
L(\theta_c, w_c, \{\bar{w}_{c'}\}_{c' \neq c}) = L_{pos}(\theta_c, w_c) + \lambda \times L_{neg} (w_c, \{\bar{w}_{c'}\}_{c' \neq c}),
\end{equation}
where $\theta_c$, $w_c$, $\{\bar{w}_{c'}\}_{c' \neq c}$, and $\lambda$ are embedding network parameters, a learnable class prototype, shared proxy class prototypes, and a scaling factor for the negative loss, respectively.
The first loss term, $L_{pos}$, is optimized to minimize the intra-class variation while the second loss term, $L_{neg}$, is optimized to maximize inter-class variation. The positive loss function is defined as follows:
\begin{equation} \label{fedhide_pos_loss}
L_{pos}(\theta_c, w_c) = (1 - w_c^T f_{\theta_c} (x))^2,
\end{equation}
where $f_{\theta_c} (x)$ is an instance embedding. Here, $f_{\theta_c} (x)$ and $w_c$ are optimized to be close each other. 
The negative loss function is defined as follows:
\begin{equation} \label{fedhide_neg_loss}
L_{neg} (w_c, \{\bar{w}_{c'}\}_{c' \neq c}) = \frac{1}{C-1} \sum_{c' \neq c } (1 + w_c^T \bar{w}_{c'})^2,
\end{equation}
where $w_c$ is optimized to be far away from the proxy class prototypes shared by the other clients.  

In FedHide, each client shares not only the updated local model but also a proxy class prototype with the server and other clients. After a local model update, a proxy class prototype is generated, as described in the next section. The proposed learning procedure is outlined in Algo. \ref{main-algo}, based on FedAvg \cite{mcmahan2017learning}.

\subsection{Proxy Class Prototype Generation by Hiding in the Neighbors} 
Our main idea is to generate proxy class prototypes that can be used in place of the true class prototypes to reduce private information leakage in the federated learning framework, where each client has access to the data of only one class.

In the proposed method, FedHide, we consider a technique for generating proxy class prototypes that conceal the true class prototype by combining it linearly with its nearest neighbors.
\begin{equation} \label{FedHide}
\bar{w}_c = \frac{\bar{w}'_c}{\| \bar{w}'_c\|_2}, ~~~ \bar{w}'_c = \alpha \cdot w_c + (1-\alpha) \cdot \frac{\sum_{c' \in \mathcal{N}^{topK}(w_c)} \bar{w}_{c'}}{\|\sum_{c' \in \mathcal{N}^{topK}(w_c)} \bar{w}_{c'}\|_2}.
\end{equation}
First, we select the top-$K$ nearest neighbors of the true class prototype. Next, we average and normalize these nearest neighbors to obtain a delegate prototype. Finally, the proxy class prototype is formed by linearly combining the true class prototype $w_c$ and the delegate prototype. In other clients within the federation, each client’s true prototype is optimized to be distant from the nearest neighbors of other clients, naturally ensuring it remains far from the true class prototypes of those clients. The values of $\alpha$ and $K$ control the amount of privacy leakage. When $\alpha=1$, $\bar{w}_c$ becomes equal to $w_c$, and the proxy class prototype is no longer private. Generally, smaller $\alpha$ and larger $K$ can reduce privacy leakage. To demonstrate efficacy more rigorously, the proposed method, FedHide, is compared with two alternative approaches.

As a comparative method, FedGN straightforwardly perturbs the true class prototype by adding random Gaussian noise,
\begin{equation} \label{FedGN}
\bar{w}_c = \frac{w_c + n}{\|w_c + n\|_2}, ~~~ n \sim \mathcal{N}(0, \sigma^2 I),
\end{equation}
where $\mathcal{N}(0, \sigma^2 I)$ is a Gaussian distribution with a zero mean and standard deviation of $\sigma$ for each element. 
It can be viewed as a mechanism of adding Gaussian random noise in differential privacy \cite{dwork2014algorithmic} except the $L^2$ normalization.
When $\sigma=0$, $\bar{w}_c$ becomes $w_c$, and the proxy class prototype is no longer private. 
Increasing $\sigma$ results in more private proxy class prototypes.

As the other comparative method, FedCS generates a proxy class prototype as follows, 
\begin{equation} \label{FedCS}
\bar{w}_c \sim \mbox{Uniform} (\{w | w^T w_c = cos(\theta), \| w \|_2 = 1 \}),
\end{equation}
where a proxy class prototype is selected uniformly at random from the vectors that exhibit a predefined cosine similarity, $cos(\theta)$, with the true class prototype $w_c$. When $cos(\theta)=1$, $\bar{w}$ becomes $w_c$, and the proxy class prototype is no longer private. By decreasing $cos(\theta)$, we can obtain more private proxy class prototypes.

\begin{figure}[t]
     \centering
     \begin{subfigure}[b]{0.45\textwidth}
         \centering
         \includegraphics[width=\textwidth]{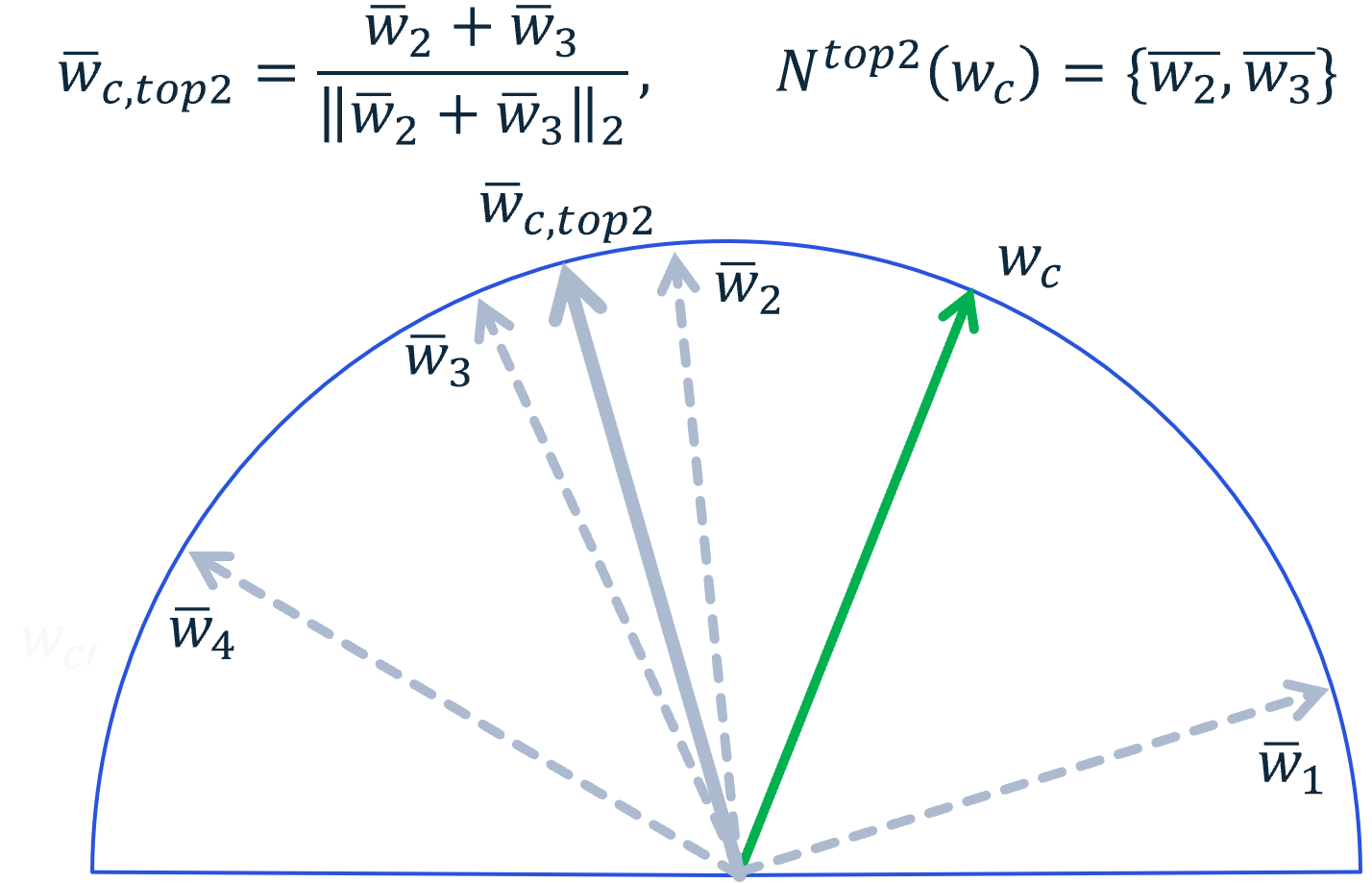}
         \caption{FedHide (find nearest neighbors)}
         \label{fig:FedHide1}
     \end{subfigure}
     \hfill
     \begin{subfigure}[b]{0.45\textwidth}
         \centering
         \includegraphics[width=\textwidth]{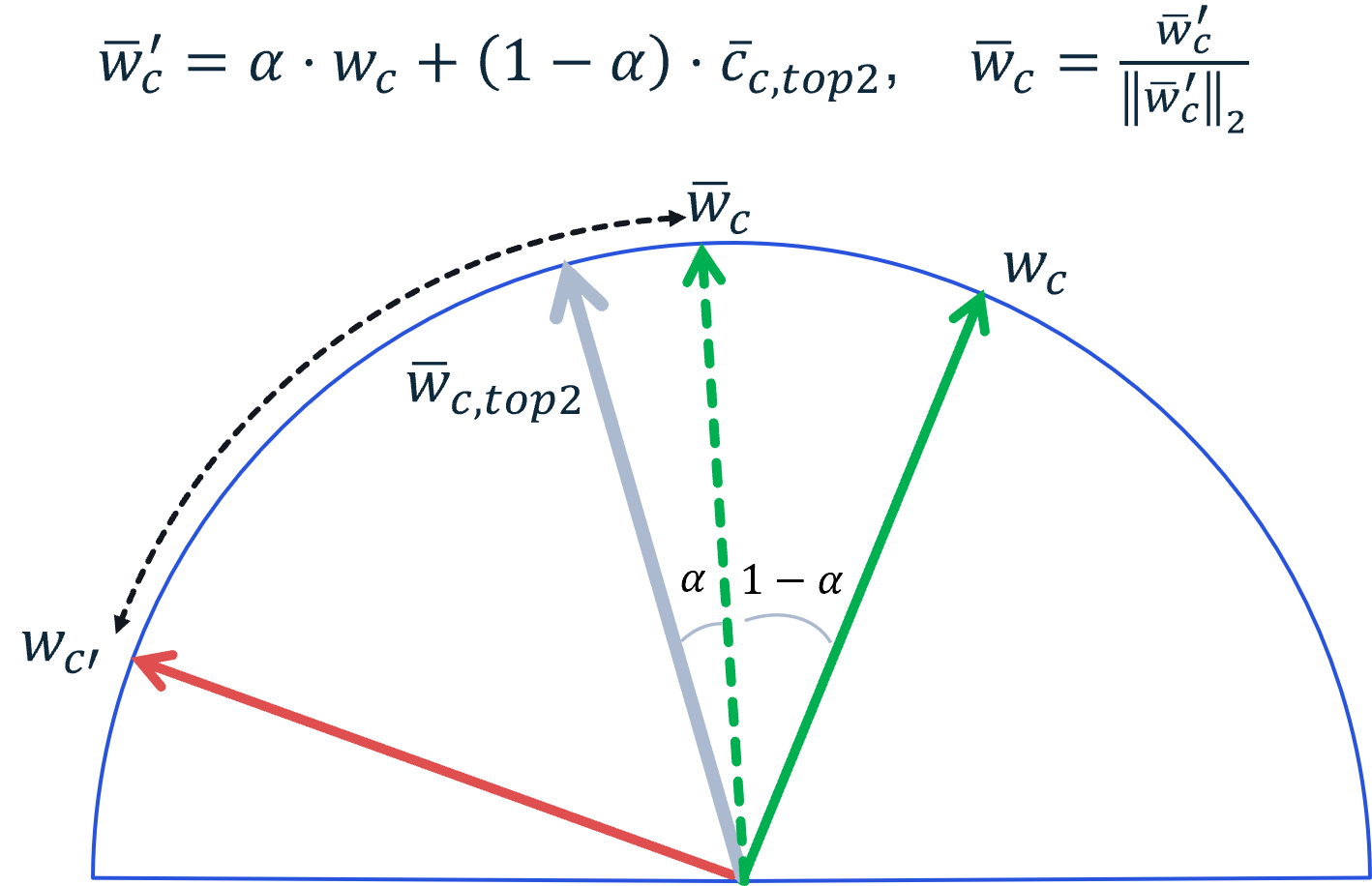}
         \caption{FedHide (linear combination)}
         \label{fig:FedHide2}
     \end{subfigure}
     
     \vspace{1.0cm}
     
     \begin{subfigure}[b]{0.45\textwidth}
         \centering
         \includegraphics[width=\textwidth]{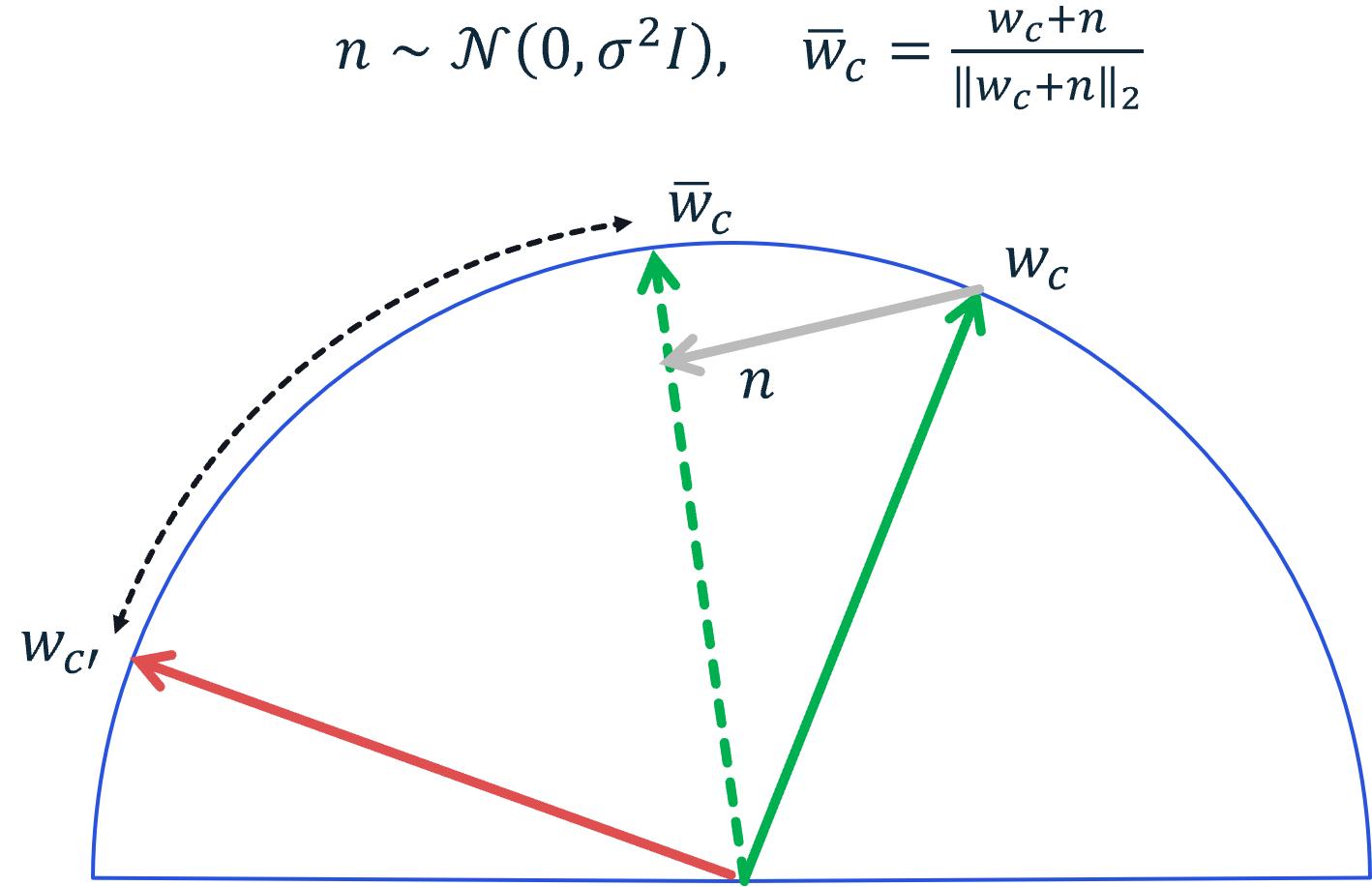}
         \caption{FedGN}
         \label{fig:FedGN}
     \end{subfigure}
     \hfill
     \begin{subfigure}[b]{0.45\textwidth}
         \centering
         \includegraphics[width=\textwidth]{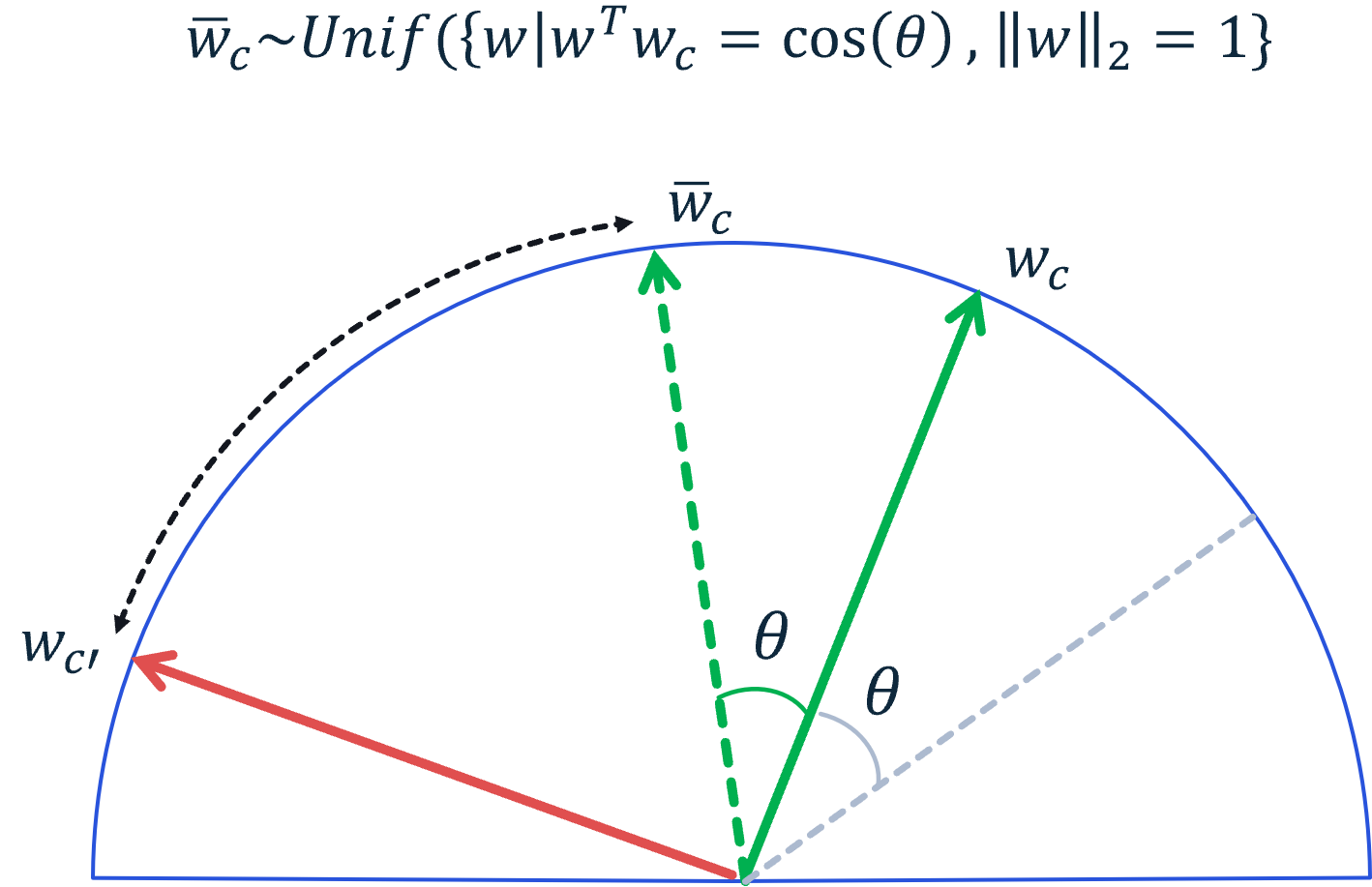}
         \caption{FedCS}
         \label{fig:FedCS}
     \end{subfigure}
    \caption{Three proxy class prototype generation methods. The negative loss is applied using the proxy class prototype $\bar{w}_c$ and other client embedding $w_{c'}$}
    \label{fedhide-figure}
\end{figure}

The proposed method (FedHide) and comparative methods (FedGN and FedCS) are illustrated in Figure \ref{fedhide-figure} and can be summarized as follows,
\begin{itemize}
\item FedHide: A linear combination of the true class prototype with an average of the top-K nearest neighbors.
\item FedGN: Addition of random Gaussian noise with zero mean and a predefined variance to the true class prototype.
\item FedCS: Random selection of a proxy class prototype based on a predefined cosine similarity with the true class prototype.
\end{itemize}

\subsection{Convergence Analysis}
We provide convergence analysis for federated embedding network learning with proxy prototypes. We assume that the following Assumptions hold for all clients $c \in \{1, 2, \ldots, C\}$, and will omit client index $c$ in the following assumptions and theorems for simplification.

\textbf{Assumption 1.}  Each local objective function is $L_1$-Lipschitz smooth, 
\begin{equation} \label{assumption_1}
\left\Vert \nabla L_{t_1} - \nabla L_{t_2}\right\Vert_2 \leq L_1 \left\Vert \phi_{t_1} - \phi_{t_2}\right\Vert_2, \forall t_1, t_2 > 0,
\end{equation}
where $\phi_{t} = \{\theta_{t}, w_{t}\}$.

\textbf{Assumption 2.}  The stochastic gradient $g_{t} = \nabla L(\phi_{t}, \xi_{t})$ is an unbiased estimator of the local gradient for each client,
\begin{equation} \label{assumption_2_1}
\mathbb{E}_{\xi \sim D}[g_{t}] = \nabla L(\phi_{t}) = \nabla L_{t}, 
\end{equation}
and its variance is bounded by $\sigma^2$,
\begin{equation} \label{assumption_2_2}
\mathbb{E}\left[ \left\Vert g_{t} - \nabla L(\phi_{t})\right\Vert_2^2 \right] \leq \sigma^2,
\end{equation}

\textbf{Assumption 3.}  The expectation of the stochastic gradient is bounded by $G_1$,
\begin{equation} \label{assumption_3}
\mathbb{E}\left[ \left\Vert g_{t} \right\Vert_2^2 \right] \leq G_1^2.
\end{equation}

\textbf{Assumption 4.}  Each local embedding function is $L_2$-Lipschitz smooth, 
\begin{equation} \label{assumption_4}
\left\Vert f(\phi_{t_1}) - f(\phi_{t_2})\right\Vert_2 \leq L_2 \left\Vert \phi_{t_1} - \phi_{t_2}\right\Vert_2, \forall t_1, t_2 > 0.
\end{equation}

\textbf{Assumption 5.}  The difference between true prototype and proxy prototype, $\delta_{t} = w_t - \bar{w}_t$ is an unbiased estimator, 
\begin{equation} \label{assumption_5_1}
\mathbb{E}_{\xi \sim D}[\delta_{t}] = \bar{\delta},
\end{equation}
and the expectation of its Euclidean norm is bounded by $G_2$, 
\begin{equation} \label{assumption_5_2}
\mathbb{E}\left[ \left\Vert \delta_{t} \right\Vert_2^2 \right] \leq G_2^2.
\end{equation}
Assumption 1 to 4 were introduced in FedProto \cite{tan2021fedproto}, and Assumption 5 is added for proxy prototype generation.
Based on the above assumptions, we present the expected loss decrease per round in the following theorem.
We adopt similar notations as in \cite{tan2021fedproto}. Specifically, we denote $e\in\{1/2, 1, 2, \ldots, E\}$ as the local iteration, $t$ as the global round, $tE$ as the time step before aggregation, and $tE + 1/2$ as the time step between aggregation and the first local model update. 

\textbf{Theorem 1.} Let Assumption 1 to 5 hold. For an arbitrary client, after every communication round, we have,
\begin{equation} \label{theorem_1}
\begin{aligned}
\mathbb{E}\left[L_{(t+1)E+1/2}\right] \leq L_{tE+1/2} - \left( \eta - \frac{L_1 \eta^2}{2}\right) \sum_{e=1/2}^{E-1} \left\Vert \nabla L_{tE+e}\right\Vert_2^2  \\
+\frac{L_1 E \eta^2}{2}\sigma^2 
+ \left( L_2^2 + \frac{\lambda}{C-1}\right)\eta^2 E G_1^2 + \frac{4\lambda}{C-1} G_2^2 
\end{aligned}
\end{equation}

\textbf{Theorem 2.} 
Let Assumption 1 to 5 hold and $\Delta = L_0 - L^*$ where $L^*$ refers to the local optimum. For an arbitrary client, given any $\epsilon > 0$, after 
\begin{equation} \label{theorem_2_1}
T = \frac{2 \Delta}{E\epsilon(2\eta - L_1 \eta^2) - E\eta^2 \left( L_1\sigma^2 + 2 \left(L_2^2 + \frac{\lambda}{C-1} \right)G_1^2   \right) - \frac{8\lambda}{C-1} G_2^2} 
\end{equation}
communication rounds with appropriate $\eta$ and $\lambda$ that ensure the denominator is positive, we have 
\begin{equation} \label{theorem_2_2}
\frac{1}{TE} \sum_{t=0}^{T-1}\sum_{e=1/2}^{E-1} \mathbb{E} \left[ \left\Vert \nabla L_{tE+e}\right\Vert_2^2 \right] < \epsilon.
\end{equation}
A detailed proof and analysis are given in Appendix.

\subsection{Prototype Leakage Measure}
Assuming that an attacker possesses a set of true class prototypes, $\mathcal{C} = \{w_1, \cdots, w_{|\mathcal{C}|}\}$, we can consider private information to be leaked when the shared proxy class prototype $\bar{w}_c$ is closest to the true class prototype $w_c$. In such cases, the input can be reconstructed using the closest class prototype. We define a prototype leakage measure as
\begin{equation} \label{PL_measure}
P_{leak} = \frac{1}{|\mathcal{C}|} \sum_{c=1}^{|\mathcal{C}|} \mathbb{1} (\mathrm{argmax}_{c'} w_{c'}^T \bar{w}_{c} = c),
\end{equation}
where $\mathbb{1}(\cdot)$ is the indicator function, which outputs $1$ when the input is true and $0$ otherwise.
It’s important to note that FedAwS consistently shares true class prototypes with the server $\bar{w}_c = w_c$, resulting in a prototype leakage of $1$.

\section{Experiments}
\subsection{Experimental Setup}

We evaluate our methods on the CIFAR-100\cite{krizhevsky2009learning}, VoxCeleb1\cite{Nagrani2017voxceleb}, and VGGFace2\cite{cao2018vggface2} datasets. Table \ref{datasets-table} provides information about these datasets.
\begin{table}[t]
  \caption{Datasets and configurations for experiments.}
  \label{datasets-table}
  \centering
  \setlength\tabcolsep{1pt} 
  \begin{tabular}{l|c|c|c}
    \toprule
                                & CIFAR-100                 & VoxCeleb1                 & VGGFace2  \\
    \midrule
    Number of clients           & $100$                       & $1211$                      & $8631$      \\
    Local data at each client   & $500$ images                & about $122$ waveforms       & about $362$ images \\
    Fraction                    & $0.1$                       & $0.01$                      & $0.001$     \\
    Model                       & ResNet-18 \cite{he2016deep} & Fast ResNet-34 \cite{chung2020defence} & MobileFaceNet \cite{chen2018mobilefacenets}   \\
    Number of rounds            & $100,000$                      & $50,000$                       & $400,000$      \\
    Validation (ACC/EER/EER)    & $100$ seen clients          & $40$ unseen clients         & $500$ unseen clients \\    
    \bottomrule
  \end{tabular}
\end{table}

\textbf{Image Classification.}
For image classification experiments, we utilize the CIFAR-100 dataset \cite{krizhevsky2009learning}, which comprises 60,000 32x32 color images across 100 classes. Our training setup involves 100 clients, each with 500 images from the same class. The remaining 100 images per class are reserved for testing. During training, we apply random horizontal flips and rotations for data augmentation. 
Our ResNet18-based embedding network is trained for 100,000 rounds, with 0.1 fraction of clients selected at each round. 
In the test phase, we calculate classification accuracy using the global model on the test set. We assume that the server has access to the clients’ test sets.

\textbf{Speaker Verification.} 
For speaker verification experiments, we utilize the VoxCeleb1 dataset \cite{Nagrani2017voxceleb}, which comprises over 100,000 utterances from 1,251 celebrities. Following the standard split, we employ 1,211 clients for training. Each client possesses approximately 122 waveforms associated with the same identity. The remaining 40 speakers are used to evaluate verification performance in terms of Equal Error Rate (EER) for 37,611 test pairs from the official test list. During the training phase, we randomly crop a 2-second temporal segment from each utterance to extract mel-scaled spectrograms using a Hamming window of 25ms length and 10ms hop size. These 40-dimensional Mel filterbanks serve as inputs to a speaker embedding network. Specifically, we use a modified Fast ResNet-34 \cite{chung2020defence} for the speaker embedding network. The embedding network is trained for 50,000 rounds, with 0.01 fraction of clients selected at each round. Notably, we do not employ any data augmentation techniques in this experiment. To calculate the EER during the test phase, we crop ten 3-second temporal segments from each utterance pair and compute the average scores across all segment pairs.

\textbf{Face Verification.} 
For face verification experiments, we utilize the VGGFace2 dataset \cite{cao2018vggface2}, which comprises 3.31 million images of 9,131 identities. Following the standard split, we employ 8,631 clients for training. Each client possesses 365 images associated with the same identity. The remaining 500 users are used to measure validation verification performance in terms of EER. We generated 338,430 test pairs to calculate EER, similar to the VoxCeleb1 evaluation. During the training phase, we first detect faces using a pretrained FaceNet \cite{schroff2015facenet}. Finally, we use 64x64 resized face images as inputs to a face embedding network. Our training process involves a MobileFaceNet-based embedding network trained for 400,000 rounds, where 0.001 fraction of clients are selected at each round.

We compare our FedHide method with FedGN, FedCS, and FedAwS. Instead of batch normalization (BN) \cite{ioffe2015batch}, we employ group normalization (GN) \cite{wu2018group} due to observations that BN does not perform well in non-iid data settings for federated learning \cite{hsieh2019non}. All models generate $L^2$-normalized 512-dimensional embedding vectors. Clients are selected in a round-robin manner, and at each client, the local model is updated with a single iteration. Across all datasets, we use a minibatch size of 16, a negative loss weight of $\lambda=10$, and a learning rate of 0.1 with the SGD optimizer. These hyperparameters were initially determined through grid search for the FedAwS experiments. Our reported results represent averages from 3 runs with different random seeds, achieved by adjusting hyperparameters. We conducted our experiments using PyTorch \cite{paszke2019pytorch} and NVIDIA RTX A5000 GPUs. For a single configuration using a GPU, the image classification, speaker verification, and face verification experiments take approximately 21 hours, 24 hours, and 80 hours, respectively.

\subsection{Results}

\begin{figure}[t]
     \centering
     \begin{subfigure}[b]{0.45\textwidth}
         \centering
         \includegraphics[width=\textwidth]{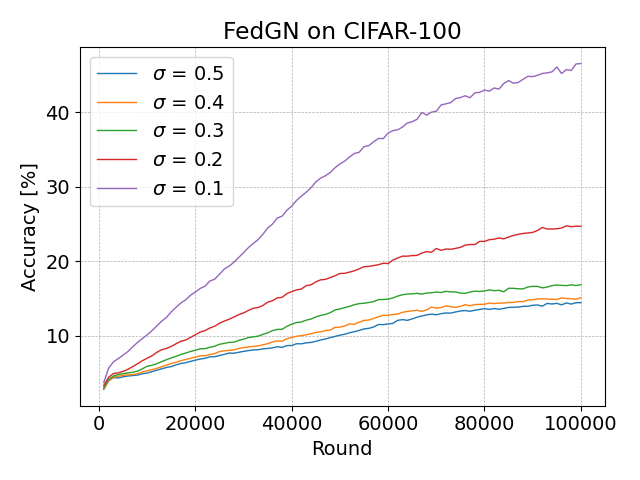}
         \caption{FedGN}
         \label{fig:cifar100_FedGN}
     \end{subfigure}
     \hfill
     \begin{subfigure}[b]{0.45\textwidth}
         \centering
         \includegraphics[width=\textwidth]{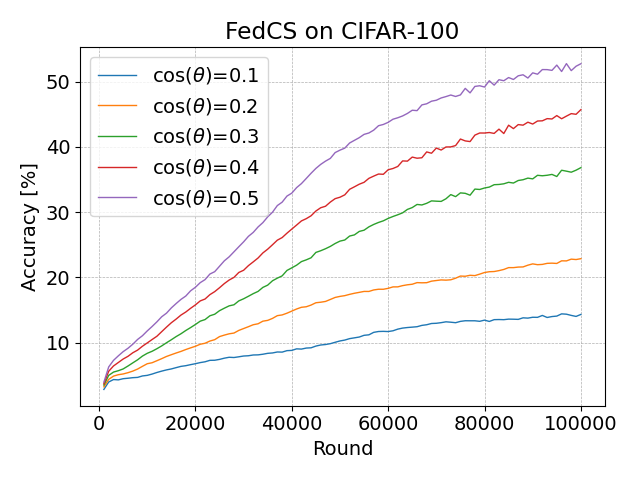}
         \caption{FedCS}
         \label{fig:cifar100_FedCS}
     \end{subfigure}
     \hfill
     
     \begin{subfigure}[b]{0.45\textwidth}
         \centering
         \includegraphics[width=\textwidth]{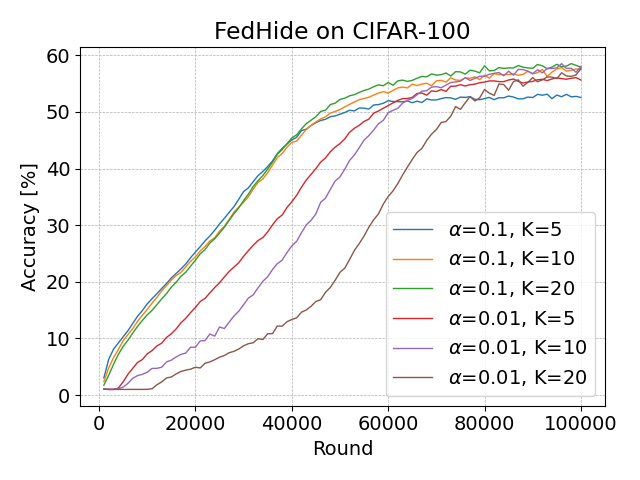}
         \caption{FedHide}
         \label{fig:cifar100_FedHide}
     \end{subfigure}
     \hfill
     \begin{subfigure}[b]{0.45\textwidth}
         \centering
         \includegraphics[width=\textwidth]{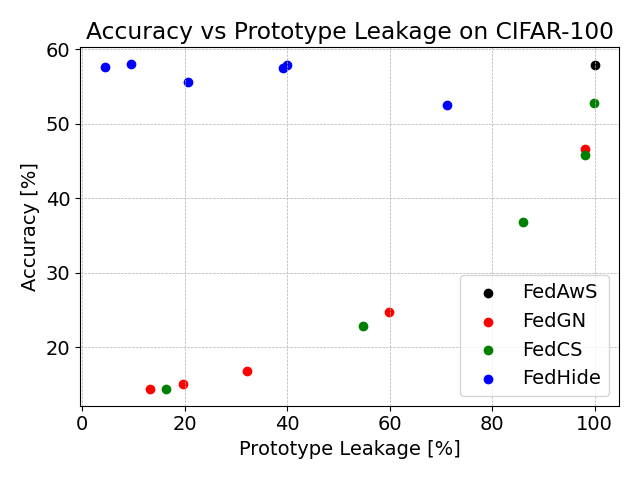}
         \caption{Accuracy versus prototype leakage}
         \label{fig:cifar100_acc_pl}
     \end{subfigure}
    \caption{FedHide results on CIFAR-100. Subfigure (a), (b), and (c) show the accuracy curves with different privacy control parameters. Subfigure (d) shows that the FedHide is the best in terms of high accuracy and low prototype leakage.}
    \label{fig:cifar100_fedhide}
\end{figure}

Figure \ref{fig:cifar100_fedhide} illustrates the experimental results of the proposed methods on the CIFAR-100 dataset. In Figure \ref{fig:cifar100_FedGN}, \ref{fig:cifar100_FedCS}, and \ref{fig:cifar100_FedHide}, the horizontal axis represents the FL round, while the vertical axis corresponds to classification accuracy. Generally, as the number of rounds increases, accuracy improves. Figure \ref{fig:cifar100_FedGN} displays the accuracy curves for FedGN with different hyperparameters, where $\sigma \in \{0.1, 0.2, 0.3, 0.4, 0.5\}$. Performance improves as $\sigma$ decreases. In Figure \ref{fig:cifar100_FedCS}, we observe the accuracy curves for FedCS with varying hyperparameters, where $cos(\theta) \in \{0.1, 0.2, 0.3, 0.4, 0.5\}$. Performance improves with increasing cosine similarities. Figure \ref{fig:cifar100_FedHide} shows the accuracy curves for FedHide with different hyperparameters, where $\alpha \in \{0.1, 0.01\}$ and $K \in \{5, 10, 20\}$. $\alpha = 0.1$ shows faster convergence than $\alpha=0.01$. Additionally, lower $K$ values result in faster convergence compared to higher values.
Figure \ref{fig:cifar100_acc_pl} presents a scatter plot for FedGN, FedCS, FedHide, and FedAwS. The horizontal axis represents prototype leakage, while the vertical axis represents accuracy at the last round. Results in the top-left corner indicate high accuracy and low prototype leakage. Notably, FedHide methods effectively reduce prototype leakage while maintaining similar accuracy of FedAwS which has high prototype leakage measure. Details can be found in Table \ref{main-results-table}. 

\begin{table}[t]
  \caption{Reconstructed images under different proxy prototype generation methods for 4 CIFAR-100 samples (S: sea, F: flower, C: chiar, P: porcupine). Lower LPIPS values indicate more privacy leakage.}
  \label{reconstructions-table}
  \centering
  \begin{tabular}{l|ccccc|ccccc}
    & \multicolumn{5}{c|}{Untrained initial model} & \multicolumn{5}{c}{Trained last model}\\
    & S & F & C & P & LPIPS & S & F & C & P & LPIPS \\
    \midrule
    Original &                                       
    \includegraphics[scale=0.4]{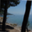} &
    \includegraphics[scale=0.4]{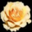} &
    \includegraphics[scale=0.4]{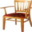} &
    \includegraphics[scale=0.4]{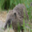} & 
     &
    \includegraphics[scale=0.4]{images/recon/sea_10_ground_truth-None.png} &
    \includegraphics[scale=0.4]{images/recon/rose_20_ground_truth-None.png} &
    \includegraphics[scale=0.4]{images/recon/chair_30_ground_truth-None.png} &
    \includegraphics[scale=0.4]{images/recon/porcupine_40_ground_truth-None.png} &  
    \\
    \midrule
    w/o perturbation &                                       
    \includegraphics[scale=0.4]{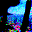} &
    \includegraphics[scale=0.4]{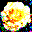} &
    \includegraphics[scale=0.4]{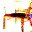} &
    \includegraphics[scale=0.4]{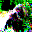} & 
    $0.48 {\scriptstyle \pm 0.05} $ &
    \includegraphics[scale=0.4]{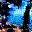} &
    \includegraphics[scale=0.4]{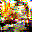} &
    \includegraphics[scale=0.4]{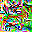} &
    \includegraphics[scale=0.4]{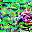} & 
    $0.64 {\scriptstyle \pm 0.07} $ \\
    \midrule
    FedGN ($0.01$) & 
    \includegraphics[scale=0.4]{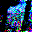} &
    \includegraphics[scale=0.4]{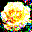} &
    \includegraphics[scale=0.4]{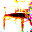} &
    \includegraphics[scale=0.4]{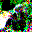} &
    $0.54 {\scriptstyle \pm 0.08} $ &
    \includegraphics[scale=0.4]{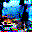} &
    \includegraphics[scale=0.4]{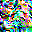} &
    \includegraphics[scale=0.4]{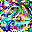} &
    \includegraphics[scale=0.4]{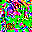} &
    $0.76 {\scriptstyle \pm 0.06} $ \\
    FedGN ($0.05$) & 
    \includegraphics[scale=0.4]{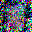} &
    \includegraphics[scale=0.4]{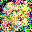} &
    \includegraphics[scale=0.4]{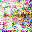} &
    \includegraphics[scale=0.4]{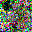} &
    $0.75 {\scriptstyle \pm 0.01} $ &
    \includegraphics[scale=0.4]{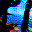} &
    \includegraphics[scale=0.4]{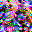} &
    \includegraphics[scale=0.4]{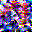} &
    \includegraphics[scale=0.4]{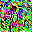} &
    $0.70 {\scriptstyle \pm 0.13} $ \\
    FedGN ($0.1$) & 
    \includegraphics[scale=0.4]{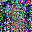} &
    \includegraphics[scale=0.4]{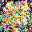} &
    \includegraphics[scale=0.4]{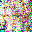} &
    \includegraphics[scale=0.4]{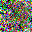} &
    $0.76 {\scriptstyle \pm 0.01} $ &
    \includegraphics[scale=0.4]{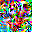} &
    \includegraphics[scale=0.4]{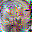} &
    \includegraphics[scale=0.4]{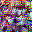} &
    \includegraphics[scale=0.4]{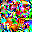} &
    $0.79 {\scriptstyle \pm 0.01} $ \\
    \midrule
    FedCS ($0.9$) &
    \includegraphics[scale=0.4]{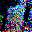} &
    \includegraphics[scale=0.4]{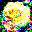} &
    \includegraphics[scale=0.4]{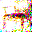} &
    \includegraphics[scale=0.4]{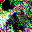} &
    $0.64 {\scriptstyle \pm 0.04} $ &
    \includegraphics[scale=0.4]{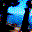} &
    \includegraphics[scale=0.4]{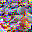} &
    \includegraphics[scale=0.4]{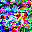} &
    \includegraphics[scale=0.4]{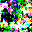} &
    $0.71 {\scriptstyle \pm 0.13} $ \\
    FedCS ($0.75$) &
    \includegraphics[scale=0.4]{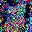} &
    \includegraphics[scale=0.4]{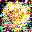} &
    \includegraphics[scale=0.4]{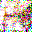} &
    \includegraphics[scale=0.4]{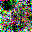} &
    $0.72 {\scriptstyle \pm 0.01} $ &
    \includegraphics[scale=0.4]{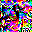} &
    \includegraphics[scale=0.4]{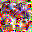} &
    \includegraphics[scale=0.4]{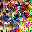} &
    \includegraphics[scale=0.4]{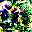} &
    $0.78 {\scriptstyle \pm 0.03} $ \\
    FedCS ($0.5$) &
    \includegraphics[scale=0.4]{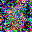} &
    \includegraphics[scale=0.4]{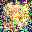} &
    \includegraphics[scale=0.4]{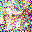} &
    \includegraphics[scale=0.4]{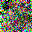} &
    $0.75 {\scriptstyle \pm 0.01} $ &
    \includegraphics[scale=0.4]{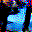} &
    \includegraphics[scale=0.4]{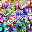} &
    \includegraphics[scale=0.4]{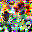} &
    \includegraphics[scale=0.4]{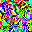} &
    $0.76 {\scriptstyle \pm 0.07} $ \\
    \midrule
    FedHide ($0.5, 1$) &
    \includegraphics[scale=0.4]{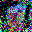} &
    \includegraphics[scale=0.4]{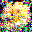} &
    \includegraphics[scale=0.4]{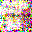} &
    \includegraphics[scale=0.4]{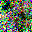} &
    $0.73 {\scriptstyle \pm 0.02} $ &
    \includegraphics[scale=0.4]{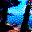} &
    \includegraphics[scale=0.4]{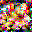} &
    \includegraphics[scale=0.4]{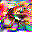} &
    \includegraphics[scale=0.4]{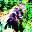} &
    $0.65 {\scriptstyle \pm 0.11} $ \\
    FedHide ($0.5, 5$) &
    \includegraphics[scale=0.4]{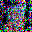} &
    \includegraphics[scale=0.4]{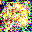} &
    \includegraphics[scale=0.4]{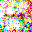} &
    \includegraphics[scale=0.4]{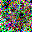} &
    $0.74 {\scriptstyle \pm 0.00} $ &
    \includegraphics[scale=0.4]{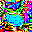} &
    \includegraphics[scale=0.4]{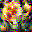} &
    \includegraphics[scale=0.4]{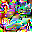} &
    \includegraphics[scale=0.4]{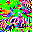} &
    $0.76 {\scriptstyle \pm 0.01} $ \\
    FedHide ($0.1, 1$) &
    \includegraphics[scale=0.4]{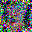} &
    \includegraphics[scale=0.4]{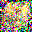} &
    \includegraphics[scale=0.4]{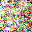} &
    \includegraphics[scale=0.4]{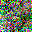} &
    $0.76 {\scriptstyle \pm 0.02} $ &
    \includegraphics[scale=0.4]{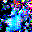} &
    \includegraphics[scale=0.4]{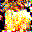} &
    \includegraphics[scale=0.4]{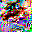} &
    \includegraphics[scale=0.4]{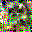} &
    $0.74 {\scriptstyle \pm 0.02} $ \\
    FedHide ($0.1, 5$) &
    \includegraphics[scale=0.4]{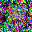} &
    \includegraphics[scale=0.4]{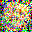} &
    \includegraphics[scale=0.4]{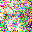} &
    \includegraphics[scale=0.4]{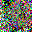} &
    $0.77 {\scriptstyle \pm 0.01} $ &
    \includegraphics[scale=0.4]{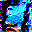} &
    \includegraphics[scale=0.4]{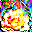} &
    \includegraphics[scale=0.4]{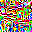} &
    \includegraphics[scale=0.4]{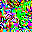} &
    $0.73 {\scriptstyle \pm 0.08} $ \\
    FedHide ($0.01, 1$) & 
    \includegraphics[scale=0.4]{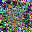} &
    \includegraphics[scale=0.4]{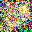} &
    \includegraphics[scale=0.4]{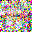} &
    \includegraphics[scale=0.4]{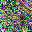} &
    $0.77 {\scriptstyle \pm 0.02} $ &
    \includegraphics[scale=0.4]{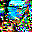} &
    \includegraphics[scale=0.4]{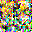} &
    \includegraphics[scale=0.4]{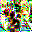} &
    \includegraphics[scale=0.4]{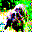} &
    $0.70 {\scriptstyle \pm 0.11} $ \\
    FedHide ($0.01, 5$) & 
    \includegraphics[scale=0.4]{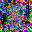} &
    \includegraphics[scale=0.4]{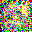} &
    \includegraphics[scale=0.4]{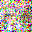} &
    \includegraphics[scale=0.4]{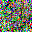} &
    $0.78 {\scriptstyle \pm 0.02} $ &
    \includegraphics[scale=0.4]{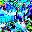} &
    \includegraphics[scale=0.4]{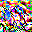} &
    \includegraphics[scale=0.4]{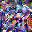} &
    \includegraphics[scale=0.4]{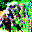} &
    $0.74 {\scriptstyle \pm 0.06} $ \\
    \bottomrule
  \end{tabular}
\end{table}

In Table \ref{reconstructions-table}, we visualize reconstructed images using a gradient inversion attack \cite{geiping2020inverting} under different proxy prototype generation methods with varying hyperparameters for four CIFAR-100 samples. For each image sample, we reconstruct the image from gradients obtained by the initial and trained ResNet-18 models using our proposed loss function (Eq. \ref{general_loss}). Additionally, we report the learned perceptual image patch similarity (LPIPS) score \cite{zhang2018unreasonable, huang2021evaluating}, where lower values indicate greater privacy leakage. We utilized official PyTorch implementations for image reconstruction from gradients \footnote{https://github.com/JonasGeiping/invertinggradients} and LPIPS scoring \footnote{https://github.com/richzhang/PerceptualSimilarity}. Notably, reconstruction using gradients from an untrained model results in higher privacy leakage (lower LPIPS) compared to reconstruction from the trained model, as trained models generally yield low-magnitude gradients. Furthermore, we observe that the hyperparameters of proxy prototype generation methods such as $\sigma, \theta, \alpha, K$ can effectively control the privacy leakage level.

Table \ref{tab:GN_CS} presents the cosine similarities between true class prototypes and proxy class prototypes $\bar{w}_c^T w_c$  in the last round of CIFAR-100 training. In the FedGN case, as $\sigma$ increases, the average cosine similarities decrease. In the FedCS case, the cosine similarity used for generating proxy class prototypes aligns naturally with the average cosine similarities. In the FedHide case, $\alpha=0.1$ shows higher cosine similarities than $\alpha=0.01$. Additionally, as $K$ increases, the average cosine similarities decrease as expected.
\begin{table}[b]
  \caption{Relations between true prototypes and proxy prototypes on CIFAR-100.}
  \label{tab:GN_CS}
  \centering
  \setlength\tabcolsep{1pt} 
  \begin{tabular}{c|ccccc|ccccc|ccc|ccc}
    \toprule
    \multirow{2}{*}{$\bar{w}_c^T w_c$}                        &  \multicolumn{5}{c|}{FedGN($\sigma$)}       & \multicolumn{5}{c|}{FedCS($cos(\theta)$)}      & \multicolumn{3}{c|}{FedHide($0.1, K$)} & \multicolumn{3}{c}{FedHide($0.01, K$)} \\ 
           & $0.1$ & $0.2$ & $0.3$ & $0.4$ & $0.5$ & $0.5$ &$0.4$ & $0.3$ & $0.2$ & $0.1$  & $5$ & $10$ & $20$ & $5$ & $10$ & $20$ \\ 
    \midrule
    AVG                    & $0.40$  & $0.22$ & $0.14$ & $0.11$ & $0.09$ & $0.50$ & $0.40$ & $0.30$ & $0.20$ & $0.10$ & $0.34$ & $0.26$ & $0.20$ & $0.23$ & $0.14$ & $0.12$ \\
    STD                     & $0.04$  & $0.04$ & $0.04$ & $0.04$ & $0.04$ & $0.00$ & $0.00$ & $0.00$ & $0.00$ & $0.00$ & $0.07$ & $0.08$ & $0.05$ & $0.10$ & $0.05$ & $0.05$  \\        
    \bottomrule
  \end{tabular}
\end{table}

\begin{figure}[t]
    \centering
    \begin{subfigure}[b]{0.45\textwidth}
        \centering
        \includegraphics[width=\textwidth]{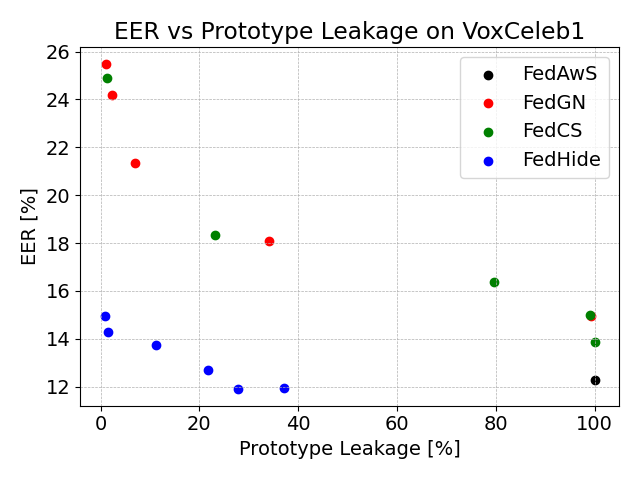}
        \caption{VoxCeleb1}
        \label{fig:voxceleb1_eer_pl}
    \end{subfigure}
    \begin{subfigure}[b]{0.45\textwidth}
        \centering
        \includegraphics[width=\textwidth]{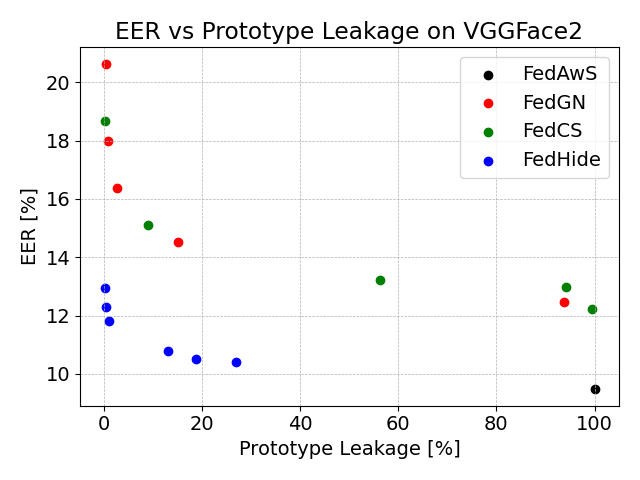}
        \caption{VGGFace2}
        \label{fig:vggface2_eer_pl}
    \end{subfigure}
    \caption{EER versus prototype leakage }
    \vspace{-0.5cm}
    \label{fig:voxceleb_vggface2_fedhide}
\end{figure}
\begingroup
\setlength{\tabcolsep}{2pt} 
\begin{table}[t]
  \caption{Overall federated learning results for the CIFAR-100, VoxCeleb1, and VGGFace2 datasets. Higher accuracy (ACC), lower equal error rate (EER), and lower prototype leakage (PL) are better.}
  \label{main-results-table}
  \centering
  \begin{tabular}{l|cc|cc|cc}
    \toprule
     & \multicolumn{2}{c|}{CIFAR-100} & \multicolumn{2}{c|}{VoxCeleb1} & \multicolumn{2}{c}{VGGFace2} \\
    \midrule
    & ACC [$\%$] & PL [$\%$] & EER [$\%$] & PL [$\%$] & EER [$\%$] & PL [$\%$] \\
    \midrule
    FedAwS                          & $57.8 {\scriptstyle \pm 2.11}$ & $100 {\scriptstyle \pm 0.00}$ & $12.3 {\scriptstyle \pm 0.41}$ & $100 {\scriptstyle \pm 0.00}$ & $9.5 {\scriptstyle \pm 0.11}$ & $100 {\scriptstyle \pm 0.00}$   \\
    \midrule
    FedGN ($0.1$)                   & $46.5 {\scriptstyle \pm 2.12}$ & $98.2 {\scriptstyle \pm 0.95}$ & $14.9 {\scriptstyle \pm 0.34}$ & $99.3 {\scriptstyle \pm 0.12}$ & $12.4 {\scriptstyle \pm 0.18}$ & $93.8 {\scriptstyle \pm 0.14}$ \\
    FedGN ($0.2$)                   & $24.7 {\scriptstyle \pm 1.49}$ & $59.9 {\scriptstyle \pm 3.27}$ & $18.1 {\scriptstyle \pm 0.32}$ & $34.1 {\scriptstyle \pm 2.07}$ & $14.5 {\scriptstyle \pm 0.19}$ & $15.1 {\scriptstyle \pm 0.11}$ \\
    FedGN ($0.3$)                   & $16.8 {\scriptstyle \pm 0.85}$ & $32.1 {\scriptstyle \pm 3.50}$ & $21.3 {\scriptstyle \pm 1.29}$ & $6.9 {\scriptstyle \pm 0.14}$ & $16.4 {\scriptstyle \pm 0.26}$ & $2.7 {\scriptstyle \pm 0.29}$ \\
    FedGN ($0.4$)                   & $15.1 {\scriptstyle \pm 0.76}$ & $19.7 {\scriptstyle \pm 2.97}$ & $24.2 {\scriptstyle \pm 1.99}$ & $2.3 {\scriptstyle \pm 0.29}$ & $18.0 {\scriptstyle \pm 0.34}$ & $0.8 {\scriptstyle \pm 0.07}$ \\
    FedGN ($0.5$)                   & $14.4 {\scriptstyle \pm 0.59}$ & $13.3 {\scriptstyle \pm 2.49}$ & $25.5 {\scriptstyle \pm 1.39}$ & $1.2 {\scriptstyle \pm 0.41}$  & $20.6 {\scriptstyle \pm 2.62}$ & $0.4 {\scriptstyle \pm 0.01}$ \\
    \midrule
    FedCS ($0.5$)                   & $52.8 {\scriptstyle \pm 2.16}$ & $100 {\scriptstyle \pm 0.00}$ & $13.9 {\scriptstyle \pm 0.56}$ & $100 {\scriptstyle \pm 0.00}$  & $12.2 {\scriptstyle \pm 0.24}$ & $99.6 {\scriptstyle \pm 0.06}$ \\
    FedCS ($0.4$)                   & $45.7 {\scriptstyle \pm 2.20}$ & $98.1 {\scriptstyle \pm 0.95}$ & $15.0 {\scriptstyle \pm 0.49}$ & $99.1 {\scriptstyle \pm 0.19}$  & $13.0 {\scriptstyle \pm 0.73}$ & $94.2 {\scriptstyle \pm 0.74}$ \\
    FedCS ($0.3$)                   & $36.8 {\scriptstyle \pm 2.19}$ & $86.0 {\scriptstyle \pm 2.43}$ & $16.4 {\scriptstyle \pm 0.33}$ & $79.6 {\scriptstyle \pm 1.42}$  & $13.2 {\scriptstyle \pm 0.19}$ & $56.2 {\scriptstyle \pm 0.78}$ \\
    FedCS ($0.2$)                   & $22.9 {\scriptstyle \pm 1.42}$ & $54.8 {\scriptstyle \pm 3.34}$ & $18.3 {\scriptstyle \pm 0.42}$ & $23.1 {\scriptstyle \pm 1.82}$ & $15.1 {\scriptstyle \pm 0.29}$ & $9.0 {\scriptstyle \pm 0.24}$ \\
    FedCS ($0.1$)                   & $14.3 {\scriptstyle \pm 0.76}$ & $16.3 {\scriptstyle \pm 2.50}$ & $24.9 {\scriptstyle \pm 1.36}$ & $1.2 {\scriptstyle \pm 0.17}$  & $18.7 {\scriptstyle \pm 0.15}$ & $0.3 {\scriptstyle \pm 0.11}$ \\
    \midrule
    FedHide ($0.1, 5/5/10$)           & $52.5 {\scriptstyle \pm 2.28}$ & $71.2 {\scriptstyle \pm 0.78}$ & $12.0 {\scriptstyle \pm 0.28}$ & $37.0 {\scriptstyle \pm 1.08}$ & $10.4 {\scriptstyle \pm 0.03}$ & $27.0 {\scriptstyle \pm 0.48}$ \\
    FedHide ($0.1, 10/10/20$)          & $57.5 {\scriptstyle \pm 2.14}$ & $39.2 {\scriptstyle \pm 3.50}$ & $11.9 {\scriptstyle \pm 0.22}$ & $27.8 {\scriptstyle \pm 0.71}$ & $10.5 {\scriptstyle \pm 0.31}$ & $18.9 {\scriptstyle \pm 0.86}$ \\
    FedHide ($0.1, 20/20/50$)          & $57.9 {\scriptstyle \pm 2.13}$ & $39.9 {\scriptstyle \pm 6.50}$ & $12.7 {\scriptstyle \pm 0.18}$ & $21.7 {\scriptstyle \pm 1.32}$ & $10.8 {\scriptstyle \pm 0.19}$ & $13.2 {\scriptstyle \pm 0.34}$ \\
    \midrule
    FedHide ($0.01, 5/5/10$)          & $55.6 {\scriptstyle \pm 2.02}$ & $20.6 {\scriptstyle \pm 1.78}$ & $13.7 {\scriptstyle \pm 0.38}$ & $11.2 {\scriptstyle \pm 2.10}$ & $11.8 {\scriptstyle \pm 0.02}$ & $1.0 {\scriptstyle \pm 0.03}$ \\
    FedHide ($0.01, 10/10/20$)         & $58.0 {\scriptstyle \pm 1.81}$ & $9.6 {\scriptstyle \pm 0.67}$ & $14.3 {\scriptstyle \pm 0.64}$ & $1.5 {\scriptstyle \pm 0.59}$  & $12.3 {\scriptstyle \pm 0.18}$ & $0.4 {\scriptstyle \pm 0.04}$ \\
    FedHide ($0.01, 20/20/50$)         & $57.6 {\scriptstyle \pm 1.59}$ & $4.3 {\scriptstyle \pm 1.04}$ & $15.0 {\scriptstyle \pm 0.64}$ & $0.8 {\scriptstyle \pm 0.04}$  & $12.9 {\scriptstyle \pm 0.02}$ & $0.2 {\scriptstyle \pm 0.02}$ \\
    \bottomrule
  \end{tabular}
\end{table}
\endgroup

Figure \ref{fig:voxceleb_vggface2_fedhide} displays scatter plots for FedAwS, FedGN, FedCS, and FedHide methods on the VoxCeleb1 and VGGFace2 datasets. The horizontal axis represents prototype leakage, while the vertical axis represents the EER at the last round. Results in the bottom-left corner indicate both low EER and low prototype leakage. Notably, the FedHide method effectively reduces prototype leakage while maintaining a similar EER to FedAwS. We use the same hyperparameters as in the CIFAR-100 experiments, except for the value of $K$ in the VGGFace2 dataset. Detailed numerical results are provided in Table \ref{main-results-table}.

\subsection{Discussion}

This paper has a few limitations. First, the FedHide method necessitates empirical hyperparameter search. We plan to explore ways to determine the prototype-dependent optimal number of nearest neighbors ($K$). Second, although we demonstrated reduced prototype leakage while maintaining accuracy empirically, we did not provide a privacy guarantee analysis. Lastly, the proposed method could be vulnerable to adaptive attackers who continuously monitor the communication channel and attempt to recover the true prototype by solving linear inverse problems \cite{lam2021gradient}.

\section{Conclusion}
We proposed FedHide, a federated learning method of embedding networks in classification and verification tasks.
In this approach, each client has access to data from only one class and cannot share a class prototype, which represents local private data, with the server or other clients. 
In the FedHide framework, clients update their local models using a contrastive learning loss to minimize intra-class variance and maximize inter-class variance. They achieved this by utilizing proxy class prototypes that can be shared among other clients.
These proxy class prototypes are generated by linearly combining them with their nearest neighbors.
In our comparative experiments, FedHide demonstrated the best performance in terms of low prototype leakage while maintaining high accuracy or low EER. Additionally, we provided a theoretical analysis of the convergence rate of FedHide when dealing with non-convex objectives.


%
%
\bibliographystyle{main}
\bibliography{main}

\title{Supplementary Material on FedHide: Federated Learning by Hiding in the Neighbors}
\author{Hyunsin Park\orcidlink{0000-0003-3556-5792} \and
Sungrack Yun\orcidlink{0000-0003-2462-3854}}
\authorrunning{H.~Park~and~S.~Yun}
\institute{Qualcomm AI Research$^{\dagger}$ \\
\email{\{hyunsinp,sungrack\}@qti.qualcomm.com}}
\maketitle

\section{Convergence Analysis for FedHide}
{\let\thefootnote\relax\footnotetext{{
$\dagger$ Qualcomm AI Research is an initiative of Qualcomm Technologies, Inc.}}}

Our convergence analysis primarily relies on \cite{tan2021fedproto, li2019convergence}. We adopt similar notations as used in \cite{tan2021fedproto}. The proposed loss function is defined as:
\begin{equation} \label{general_loss_appendix}
L = (1 - w_c^T f_{\theta_c})^2 + \frac{\lambda}{C-1} \sum_{c' \neq c } (1 + w_c^T \bar{w}_{c'})^2.
\end{equation}
Regarding iteration notations, we use $t$ for global communication rounds and $e \in \{1/2, 1, 2, \ldots, E\}$ for local iterations. Additionally, $tE$ represents the time step before aggregation at the server, and $tE+1/2$ denotes the initial time step before the first local iteration.

\subsubsection{Assumptions} \hfill

\textbf{Assumption 1.}  Each local objective function is $L_1$-Lipschitz smooth, 
\begin{equation} \label{assumption_1}
\left\Vert \nabla L_{t_1} - \nabla L_{t_2}\right\Vert_2 \leq L_1 \left\Vert \phi_{t_1} - \phi_{t_2}\right\Vert_2, \forall t_1, t_2 > 0,
\end{equation}
where $\phi_{t} = \{\theta_{t}, w_{t}\}$.

\textbf{Assumption 2.}  The stochastic gradient $g_{t} = \nabla L(\phi_{t}, \xi_{t})$ is an unbiased estimator of the local gradient for each client,
\begin{equation} \label{assumption_2_1}
\mathbb{E}_{\xi \sim D}[g_{t}] = \nabla L(\phi_{t}) = \nabla L_{t}, 
\end{equation}
and its variance is bounded by $\sigma^2$,
\begin{equation} \label{assumption_2_2}
\mathbb{E}\left[ \left\Vert g_{t} - \nabla L(\phi_{t})\right\Vert_2^2 \right] \leq \sigma^2,
\end{equation}

\textbf{Assumption 3.}  The expectation of the stochastic gradient is bounded by $G_1$,
\begin{equation} \label{assumption_3}
\mathbb{E}\left[ \left\Vert g_{t} \right\Vert_2^2 \right] \leq G_1^2.
\end{equation}

\textbf{Assumption 4.}  Each local embedding function is $L_2$-Lipschitz smooth, 
\begin{equation} \label{assumption_4}
\left\Vert f(\phi_{t_1}) - f(\phi_{t_2})\right\Vert_2 \leq L_2 \left\Vert \phi_{t_1} - \phi_{t_2}\right\Vert_2, \forall t_1, t_2 > 0.
\end{equation}

\textbf{Assumption 5.}  The difference between true prototype and proxy prototype, $\delta_{t} = w_t - \bar{w}_t$ is an unbiased estimator, 
\begin{equation} \label{assumption_5_1}
\mathbb{E}_{\xi \sim D}[\delta_{t}] = \bar{\delta},
\end{equation}
and the expectation of its Euclidean norm is bounded by $G_2$, 
\begin{equation} \label{assumption_5_2}
\mathbb{E}\left[ \left\Vert \delta_{t} \right\Vert_2^2 \right] \leq G_2^2.
\end{equation}

\subsubsection{Key Lemmas} \hfill

\textbf{Lemma 1.} Let Assumption 1 and 2 hold. From the beginning of communication of round $t+1$ to the last local update step, the loss function of an arbitrary client can be bounded as:
\begin{equation}
\mathbb{E}\left[L_{(t+1)E}\right] \leq L_{tE+1/2} - \left( \eta - \frac{L_1 \eta^2}{2}\right) \sum_{e=1/2}^{E-1} \left\Vert \nabla L_{tE+e}\right\Vert_2^2 +\frac{L_1 E \eta^2}{2}\sigma^2.
\end{equation}
This Lemma is same with the Lemma 1 in \cite{tan2021fedproto}. Please refer the detailed proof in the paper.

\textbf{Lemma 2.} Let Assumption 3, 4, and 5 hold. After the model and prototype aggregation at the server, the loss function of an arbitrary client can be bounded as:
\begin{equation}
\mathbb{E}\left[L_{(t+1)E +1/2}\right] \leq L_{(t+1)E} + \left( L_2^2 + \frac{\lambda}{C-1}\right)\eta^2 E G_1^2 + \frac{4\lambda}{C-1} G_2^2.
\end{equation}

\begin{proof}
\begin{align}
   & L_{(t+1)E + 1/2} \nonumber \\ 
   &= L_{(t+1)E} + L_{(t+1)E + 1/2} - L_{(t+1)E} \\
   &= L_{(t+1)E} + \left( 1 - w_{c,(t+1)E}^T f_{\theta, (t+1)E + 1/2}\right)^2  \nonumber \\
   & ~~~~~~~~~~~~~~ - \left( 1 - w_{c,(t+1)E}^T f_{\theta, (t+1)E}\right)^2 \nonumber \\
   & ~~~~~~~~~~~~~~ + \frac{\lambda}{C-1}\sum_{c' \neq c} \left( 1 + w_{c,(t+1)E}^T \bar{w}_{c',(t+1)E + 1/2}  \right)^2 \nonumber \\
   & ~~~~~~~~~~~~~~ - \frac{\lambda}{C-1}\sum_{c' \neq c} \left( 1 + w_{c,(t+1)E}^T \bar{w}_{c',(t+1)E}  \right)^2.
\end{align}
Here, let $A$ be
\begin{align}
& \left( 1 - w_{c,(t+1)E}^T f_{\theta, (t+1)E + 1/2}\right)^2 - \left( 1 - w_{c,(t+1)E}^T f_{\theta, (t+1)E}\right)^2 \\
& \overset{(a)}{\leq} \left( w_{c,(t+1)E}^T \left( f_{\theta, (t+1)E + 1/2} - f_{\theta, (t+1)E} \right)\right)^2 \\
& \overset{(b)}{\leq} \left( \lVert w_{c,(t+1)E} \rVert_2 \lVert f_{\theta, (t+1)E + 1/2} - f_{\theta, (t+1)E} \rVert_2 \right)^2 \\
& \overset{(c)}{=} \lVert f_{\theta, (t+1)E + 1/2} - f_{\theta, (t+1)E} \rVert_2^2 \\
& \overset{(d)}{\leq} L_2^2 \lVert \theta_{(t+1)E + 1/2} - \theta_{(t+1)E} \rVert_2^2 \\
& \overset{(e)}{\leq} L_2^2 \lVert \phi_{(t+1)E + 1/2} - \phi_{(t+1)E} \rVert_2^2 \\
& =  L_2^2 \lVert (\phi_{(t+1)E} - \phi_{tE + 1/2}) - (\phi_{(t+1)E + 1/2} - \phi_{tE + 1/2}) \rVert_2^2.
\end{align}
Take expectation of random variable $\xi$, then
\begin{align}
\mathbb{E}[A] & \overset{(f)}{\leq} L_2^2 \mathbb{E} \lVert \phi_{(t+1)E} - \phi_{tE + 1/2} \rVert_2^2 \\
& = L_2^2 \eta^2 \mathbb{E} \lVert \sum_{e=1/2}^{E-1} g_{tE+e}\rVert_2^2 \\
& \overset{(a)}{\leq} L_2^2 \eta^2   \sum_{e=1/2}^{E-1} \lVert \mathbb{E} g_{tE+e} \rVert_2^2 \\
& \overset{(g)}{\leq} L_2^2 \eta^2 EG_1^2 
\end{align}
And, let $B$ be
\begin{align}
& \left( 1 + w_{c,(t+1)E}^T \bar{w}_{c',(t+1)E + 1/2}  \right)^2 -  \left( 1 + w_{c,(t+1)E}^T \bar{w}_{c',(t+1)E}  \right)^2 \\
& \overset{(a)}{\leq} \left( w_{c,(t+1)E}^T \left( \bar{w}_{c',(t+1)E + 1/2}  - \bar{w}_{c',(t+1)E}  \right) \right)^2 \\
& \overset{(b)}{\leq} \left( \lVert w_{c,(t+1)E} \rVert_2 \lVert \bar{w}_{c',(t+1)E + 1/2}  -  \bar{w}_{c',(t+1)E}  \rVert_2 \right)^2 \\
& \overset{(c)}{=} \lVert \bar{w}_{c',(t+1)E + 1/2}  - \bar{w}_{c',(t+1)E}  \rVert_2^2 \\
& = \lVert (w_{c',(t+1)E + 1/2} - w_{c',(t+1)E}) - (\delta_{c',(t+1)E + 1/2} - \delta_{c',(t+1)E})   \rVert_2^2 \\
& = \lVert w_{c',(t+1)E + 1/2} - w_{c',(t+1)E}) \rVert_2^2  \\
& ~~~~ -2 (w_{c',(t+1)E + 1/2} - w_{c',(t+1)E})^T(\delta_{c',(t+1)E + 1/2} - \delta_{c',(t+1)E}) \\
& ~~~~ + \lVert \delta_{c',(t+1)E + 1/2} - \delta_{c',(t+1)E} \rVert_2^2
\end{align}
Take expectation of random variable $\xi$, then
\begin{align}
\mathbb{E}[B] & \overset{(h)}{\leq} \eta^2EG_1^2  + \mathbb{E} [\lVert \delta_{c',(t+1)E + 1/2} - \delta_{c',(t+1)E} \rVert_2^2 ]\\
& \overset{(b)}{\leq} \eta^2EG_1^2 + \mathbb{E} [\lVert \delta_{c',(t+1)E + 1/2} \rVert_2^2]  + \mathbb{E}[\lVert \delta_{c',(t+1)E} \rVert_2^2 ] \\
& ~~~~~~~~~~~~~~ + 2 \mathbb{E}[\lVert \delta_{c',(t+1)E + 1/2} \rVert_2 \lVert \delta_{c',(t+1)E} \rVert_2] \\
& \overset{(i)}{\leq} \eta^2EG_1^2 + 4G_2^2
\end{align}
Lastly, by taking expectation of random variable $\xi$ on Eq. (12) and based on $\mathbb{E}[A]$ and $\mathbb{E}[B]$, we obtain
\begin{equation}
\mathbb{E}\left[L_{(t+1)E +1/2}\right] \leq L_{tE+1} + \left( L_2^2 + \frac{\lambda}{C-1}\right)\eta^2 E G_1^2 + \frac{4\lambda}{C-1} G_2^2.
\end{equation}
In the above derivations, $(a)$ follows from Jensen's inequality, $(b)$ follows from Cauchy–Schwarz inequality, $(c)$ follows from the $l2$-normalized prototypes, $(d)$ follows from $L_2$-Lipschitz continuity in Assumption 4, $(e)$ follows from the fact that $\theta$ is a subset of $\phi$, $(f)$ follows from $\mathbb{E}\lVert X-\mathbb{E}[X] \rVert_2 ^2 \leq \mathbb{E}\lVert X \rVert_2^2$, $(g)$ follows from Assumption 3, $(h)$ follows from Eq. (23) and Assumption 5, $(i)$ follows from Assumption 5.\hfill$\square$
\end{proof}

\subsubsection{Theorems} \hfill

\textbf{Theorem 1.} Let Assumption 1 to 5 hold. For an arbitrary client, after every communication round, we have,
\begin{equation} \label{theorem_1}
\begin{aligned}
\mathbb{E}\left[L_{(t+1)E+1/2}\right] \leq L_{tE+1/2} - \left( \eta - \frac{L_1 \eta^2}{2}\right) \sum_{e=1/2}^{E-1} \left\Vert \nabla L_{tE+e}\right\Vert_2^2  \\
+\frac{L_1 E \eta^2}{2}\sigma^2 
+ \left( L_2^2 + \frac{\lambda}{C-1}\right)\eta^2 E G_1^2 + \frac{4\lambda}{C-1} G_2^2 
\end{aligned}
\end{equation}

\begin{proof}
Taking expectation on both sides in Lemma 1 and 2, then sum them, we can easily obtain the Theorem 1. 
\hfill$\square$
\end{proof}
\textbf{Theorem 2.} 
Let Assumption 1 to 5 hold and $\Delta = L_0 - L^*$ where $L^*$ refers to the local optimum. For an arbitrary client, given any $\epsilon > 0$, after 
\begin{equation} \label{theorem_2_1}
T = \frac{2 \Delta}{E\epsilon(2\eta - L_1 \eta^2) - E\eta^2 \left( L_1\sigma^2 + 2 \left(L_2^2 + \frac{\lambda}{C-1} \right)G_1^2   \right) - \frac{8\lambda}{C-1} G_2^2} 
\end{equation}
communication rounds with appropriate $\eta$ and $\lambda$ that ensure the denominator is positive, we have 
\begin{equation} \label{theorem_2_2}
\frac{1}{TE} \sum_{t=0}^{T-1}\sum_{e=1/2}^{E-1} \mathbb{E} \left[ \left\Vert \nabla L_{tE+e}\right\Vert_2^2 \right] < \epsilon.
\end{equation}

\begin{proof}
Take expectation on both sides in Eq. (\ref{theorem_1}), then telescope considering the communication round from $t=0$ to $t=T-1$ with the time step from $e=1/2$ to $e=E-1$ in each communication round, we have
\begin{align}
&\frac{1}{TE} \sum_{t=0}^{T-1}\sum_{e=1/2}^{E-1} \mathbb{E} \left[ \left\Vert \nabla L_{tE+e}\right\Vert_2^2 \right] \\
&\leq \left[ \frac{1}{TE}\sum_{t=0}^{T-1}(L_{tE+1/2} - \mathbb{E}[L_{(t+1)E+1/2}]) +\frac{L_1 E \eta^2}{2}\sigma^2 \right. \nonumber \\ 
&~~ \left.  + \left( L_2^2 + \frac{\lambda}{C-1}\right)\eta^2 E G_1^2 + \frac{4\lambda}{C-1} G_2^2 \right] /  (\eta - L_1\eta^2/2).
\end{align}
Given any $\epsilon > 0$, let
\begin{align}
&\left[ \frac{1}{TE}\sum_{t=0}^{T-1}(L_{tE+1/2} - \mathbb{E}[L_{(t+1)E+1/2}]) +\frac{L_1 E \eta^2}{2}\sigma^2 \right. \nonumber \\ 
&~~ \left.  + \left( L_2^2 + \frac{\lambda}{C-1}\right)\eta^2 E G_1^2 + \frac{4\lambda}{C-1} G_2^2 \right] /  (\eta - L_1\eta^2/2) < \epsilon,
\end{align}
that is
\begin{align}
&\left[ \frac{2}{TE}\sum_{t=0}^{T-1}(L_{tE+1/2} - \mathbb{E}[L_{(t+1)E+1/2}]) + L_1 E \eta^2\sigma^2 \right. \nonumber \\ 
&~~ \left.  + 2\left( L_2^2 + \frac{\lambda}{C-1}\right)\eta^2 E G_1^2 + \frac{8\lambda}{C-1} G_2^2 \right] /  (2\eta - L_1\eta^2) < \epsilon.
\end{align}
Let $\Delta = L_0 - L^*$, Since $\sum_{t=0}^{T-1}(L_{tE+1/2} - \mathbb{E}[L_{(t+1)E+1/2}]) \leq \Delta$, the above equation holds when
\begin{align}
\frac{ \frac{2\Delta}{TE} + L_1 E \eta^2\sigma^2 + 2\left( L_2^2 + \frac{\lambda}{C-1}\right)\eta^2 E G_1^2 + \frac{8\lambda}{C-1} G_2^2}{2\eta - L_1\eta^2}
 < \epsilon.
\end{align}
That is 
\begin{equation}
T = \frac{2 \Delta}{E\epsilon(2\eta - L_1 \eta^2) - E\eta^2 \left( L_1\sigma^2 + 2 \left(L_2^2 + \frac{\lambda}{C-1} \right)G_1^2   \right) - \frac{8\lambda}{C-1} G_2^2}.
\end{equation}
So, we have 
\begin{equation}
\frac{1}{TE} \sum_{t=0}^{T-1}\sum_{e=1/2}^{E-1} \mathbb{E} \left[ \left\Vert \nabla L_{tE+e}\right\Vert_2^2 \right] < \epsilon,
\end{equation}
when $\eta$ and $\lambda$ are set to ensure the denominator of Eq. (45). 
\hfill$\square$
\end{proof}

\section{Visualization of the Learned Embedding Spaces}
\begin{figure}[ht]
     \centering
     \begin{subfigure}[b]{0.48\textwidth}
         \centering
         \includegraphics[width=\textwidth]{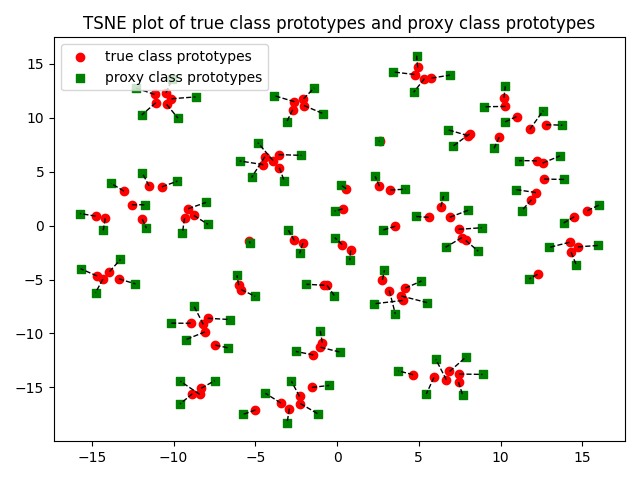}
         \caption{$\sigma=0.1,~\mbox{ACC}= 46.5\%,~\mbox{PL}=98.2\%$}
         \label{fig:fedhidegn_0p1}
     \end{subfigure}
     \hfill
     \begin{subfigure}[b]{0.48\textwidth}
         \centering
         \includegraphics[width=\textwidth]{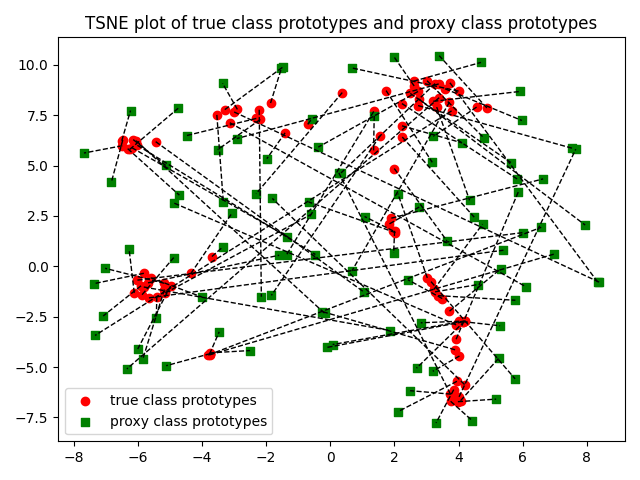}
         \caption{$\sigma=0.5,~\mbox{ACC}= 14.4\%,~\mbox{PL}=13.3\%$}
         \label{fig:fedhidegn_0p5}
     \end{subfigure}
    \caption{t-SNE visualizations of FedGN methods for the CIFAR-100 dataset. (red circle: true class prototype, green square: proxy class prototype, dashed line: pairs of true and proxy prototypes)}
    \label{fedhidegn_tsne}
\end{figure}
Figure \ref{fedhidegn_tsne} shows t-SNE (t-distributed stochastic neighbor embedding) \cite{van2008visualizing} visualizations of FedGN methods ($\sigma \in \{0.1, 0.5\}$) for the CIFAR-100 dataset, where cosine similarity was used for t-SNE metric. As shown in the figure, we can observe that FedGN with $\sigma=0.1$ shows closer distances between the true class prototypes and the corresponding proxy class prototypes than FedGN with $\sigma=0.5$. 
Compared with FedGN with $\sigma=0.1$, FedGN with $\sigma=0.5$ learns true class prototypes that are not separated well, so it shows low accuracy of $14.4\%$ although it gives low prototype leakage of $13.3\%$.
 
\begin{figure}[ht]
    \vspace{0.5cm}
     \begin{subfigure}[b]{0.48\textwidth}
         \centering
         \includegraphics[width=\textwidth]{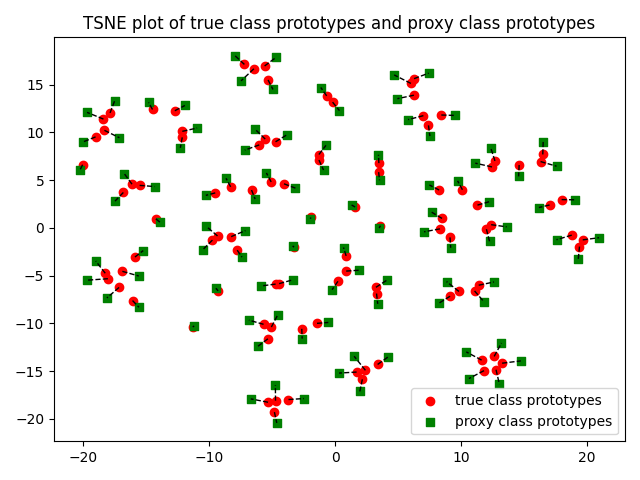}
         \caption{$cos(\theta)=0.5,~\mbox{ACC} = 52.8\%,~ \mbox{PL} = 100\%$}
         \label{fig:fedhidecs_0p1}
     \end{subfigure}
     \hfill
     \begin{subfigure}[b]{0.48\textwidth}
         \centering
         \includegraphics[width=\textwidth]{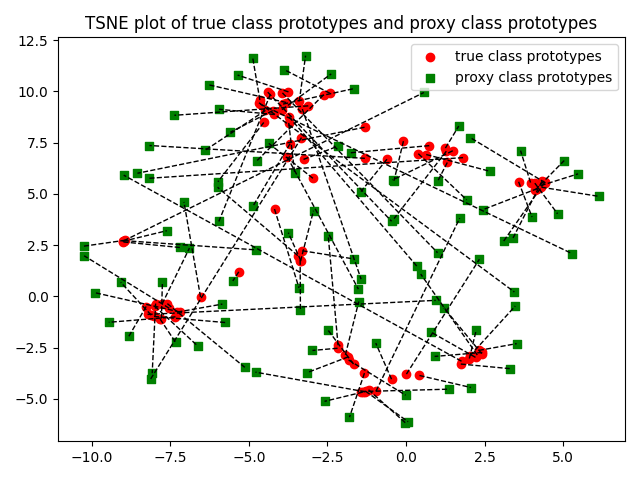}
         \caption{$cos(\theta)=0.1,~\mbox{ACC} = 14.3\%,~ \mbox{PL} = 16.3\%$}
         \label{fig:fedhidecs_0p5}
     \end{subfigure}
    \caption{t-SNE visualizations of FedCS methods for the CIFAR-100 dataset. (red circle: true class prototype, green square: proxy class prototype, dashed line: pairs of true and proxy class prototypes)}
    \label{fedhidecs_tsne}
\end{figure}
Figure \ref{fedhidecs_tsne} depicts t-SNE visualizations of FedCS methods ($cos(\theta) \in \{0.5, 0.1\}$) for the CIFAR-100 dataset. Similarly with the results of FedGN, we can observe that FedCS with $cos(\theta)=0.5$ shows closer distances between the true class prototypes and the corresponding proxy class prototypes than FedCS with $cos(\theta)=0.1$. 
Compared with FedCS with $cos(\theta)=0.5$, FedCS with $cos(\theta)=0.1$ learns true class prototypes that are not separated well, so it shows low accuracy of $14.3\%$ although it gives low prototype leakage of $16.3\%$.

\begin{figure}[ht]
    \vspace{0.5cm}
     \begin{subfigure}[b]{0.48\textwidth}
         \centering
         \includegraphics[width=\textwidth]{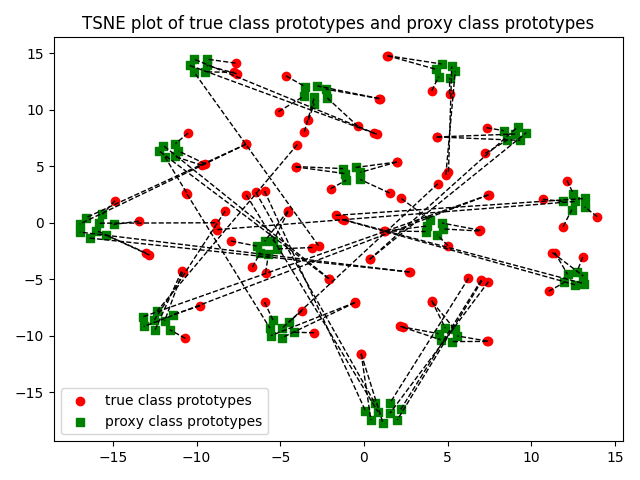}
         \caption{$\alpha=0.1, K=5, \mbox{ACC} = 52.5\%, \mbox{PL} = 71.2\%$}
         \label{fig:fedhidenn_0p1_nh5}
     \end{subfigure}
     \hfill
     \begin{subfigure}[b]{0.48\textwidth}
         \centering
         \includegraphics[width=\textwidth]{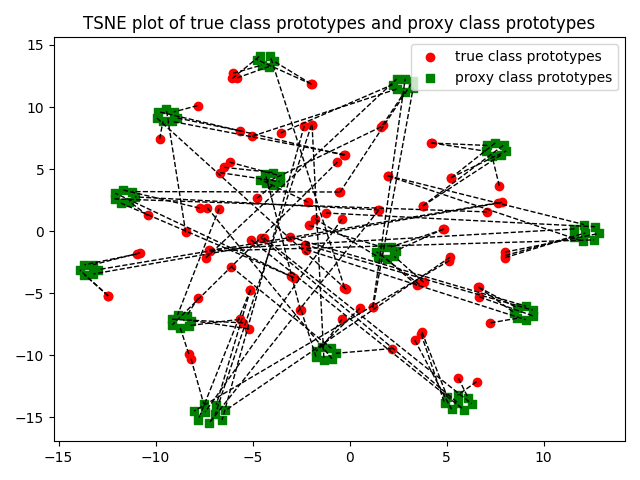}
         \caption{$\alpha=0.01, K=5, \mbox{ACC} = 55.6\%, \mbox{PL} = 20.6\%$}
         \label{fig:fedhidenn_0p01_nh5}
     \end{subfigure}
     \hfill
     \begin{subfigure}[b]{0.48\textwidth}
         \centering
         \includegraphics[width=\textwidth]{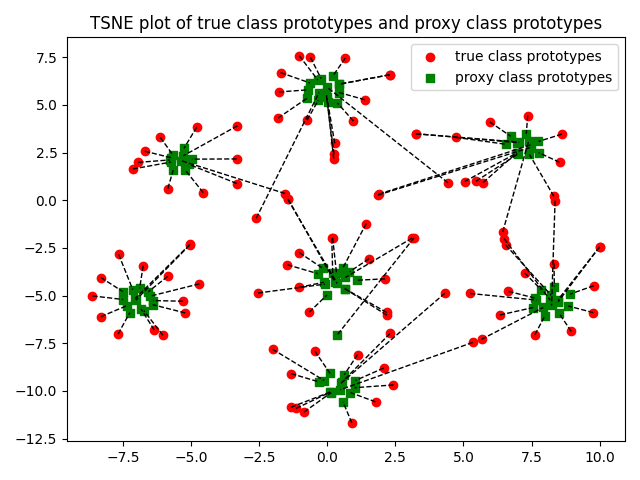}
         \caption{$\alpha=0.1, K=10, \mbox{ACC} = 57.5\%, \mbox{PL} = 39.2\%$}
         \label{fig:fedhidenn_0p1_nh10}
     \end{subfigure}
     \hfill
     \begin{subfigure}[b]{0.48\textwidth}
         \centering
         \includegraphics[width=\textwidth]{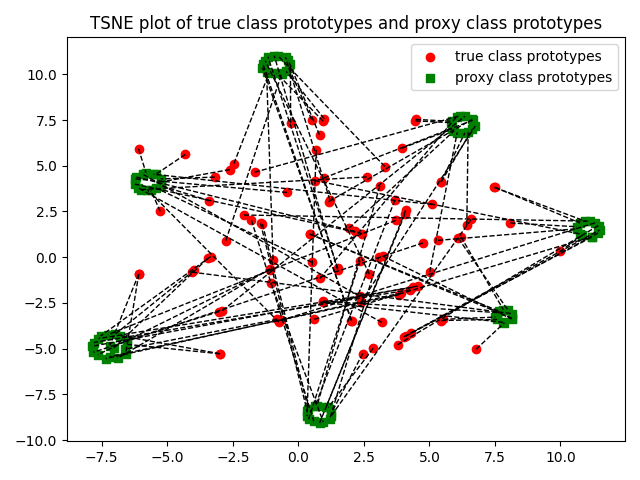}
         \caption{$\alpha=0.01, K=10, \mbox{ACC} = 58.0\%, \mbox{PL} = 9.6\%$}
         \label{fig:fedhidenn_0p01_nh10}
     \end{subfigure}
     \hfill
     \begin{subfigure}[b]{0.48\textwidth}
         \centering
         \includegraphics[width=\textwidth]{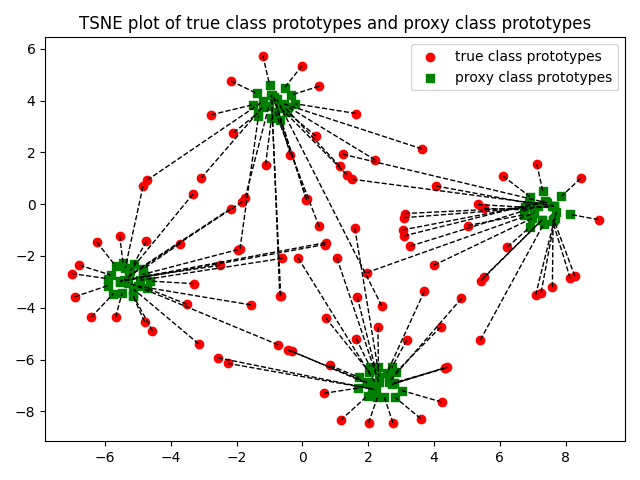}
         \caption{$\alpha=0.1, K=20, \mbox{ACC} = 57.9\%, \mbox{PL} = 39.9\%$}
         \label{fig:fedhidenn_0p1_nh20}
     \end{subfigure}
     \hfill
     \begin{subfigure}[b]{0.48\textwidth}
         \centering
         \includegraphics[width=\textwidth]{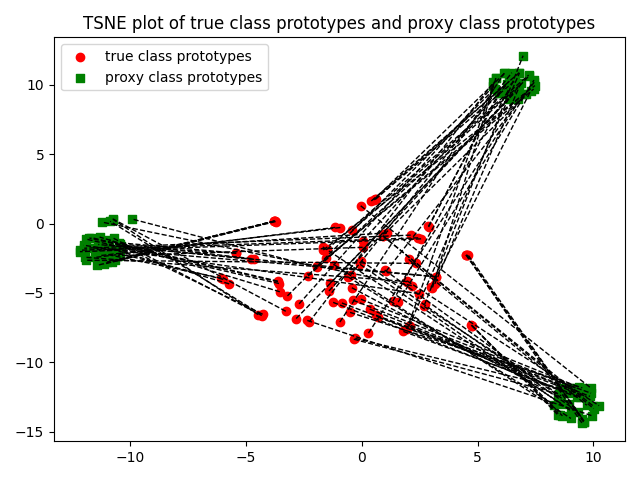}
         \caption{$\alpha=0.01, K=20, \mbox{ACC} = 57.6\%, \mbox{PL} = 4.3\%$}
         \label{fig:fedhidenn_0p01_nh20}
     \end{subfigure}
     \hfill
    \caption{t-SNE visualizations of FedHide methods for the CIFAR-100 dataset. (red circle: true class prototype, green square: proxy class prototype, dashed line: pairs of true and proxy class prototypes)}
    \vspace{-0.3cm}
    \label{fedhidenn_tsne}
\end{figure}
Figure \ref{fedhidenn_tsne} presents t-SNE visualizations of FedHide methods ($\alpha \in \{0.1, 0.01\}$, $K \in \{5, 10, 20\}$) for the CIFAR-100 dataset.
As shown in the figure, due the nature of FedHide algorithm, proxy class prototypes are more grouped compared with FedGN and FedCS. Note that the proxy class prototypes hide in not the true class prototypes but the other proxy class prototypes.   
We can observe that FedHide with $\alpha = 0.01$ shows farther distances between the true class prototype and the corresponding proxy class prototypes than FedHide with $\alpha = 0.1$.
Since there are more overlaps between nearest neighbor sets of clients as $K$ increases, the proxy prototypes calculated with these nearest neighbors would be closer and grouped.

\end{document}